\documentclass[lettersize,journal]{IEEEtran}
\usepackage{amsmath,amsfonts}
\usepackage[linesnumbered,ruled,vlined]{algorithm2e}
\usepackage{array}
\usepackage[caption=false,font=normalsize,labelfont=sf,textfont=sf]{subfig}
\usepackage{textcomp}
\usepackage{stfloats}
\usepackage{url}
\usepackage{verbatim}
\usepackage{graphicx}
\usepackage[style=ieee]{biblatex}
\usepackage{booktabs}
\usepackage{multirow}
\usepackage{makecell}
\usepackage{longtable}
\usepackage{threeparttable}
\usepackage{caption}
\usepackage{tikz}
\usepackage{amsthm}
\newtheorem{theorem}{\textbf{Theorem}}
\newtheorem{lemma}{Lemma}
\theoremstyle{definition}
\newtheorem{remark}{Remark}

\usepackage{xcolor}

\begin{document}

\title{Data-Aware Gradient Compression for FL in Communication-Constrained Mobile Computing}

\author{Rongwei Lu, \textit{Student Member, IEEE}, Yutong Jiang, Yinan Mao, \textit{Student Member, IEEE}, Chen Tang, \\Bin Chen, \textit{Member, IEEE}, Laizhong Cui, \textit{Senior Member, IEEE}, Zhi Wang, \textit{Senior Member, IEEE}

\thanks{Rongwei Lu, Yutong Jiang and Zhi Wang are with Shenzhen International Graduate School, Tsinghua University, Shenzhen 518000, China (e-mail: \{lurw24, jiang-yt24, wangzhi\}@mails.tsinghua.edu.cn).}
\thanks{Yinan Mao is with the AI Cloud Business Group, Baidu Incorporated, Beijing 100085, China (e-mail: maoyinan01@baidu.com).}
\thanks{Chen Tang is with the Department of Computer Science and Technology, Tsinghua University, Beijing 100084, China (e-mail: genprtung@gmail.com).}
\thanks{Bin Chen is with the Department of Computer Science and Technology, Harbin Institute of Technology, Shenzhen 518055, China (e-mail: chenbin2021@hit.edu.cn).}
\thanks{Laizhong Cui is with the College of Computer Science and
Software Engineering, Shenzhen University, Shenzhen 518060, China (e-mail: cuilz@szu.edu.cn).}
}

\markboth{}
{Shell \MakeLowercase{\textit{et al.}}: A Sample Article Using IEEEtran.cls for IEEE Journals}

\IEEEpubid{}

\maketitle
\begin{abstract}

Federated Learning (FL) in mobile environments faces significant communication bottlenecks. Gradient compression has proven as an effective solution to this issue, offering substantial benefits in environments with limited bandwidth and metered data. Yet, it encounters severe performance drops in non-IID environments due to a one-size-fits-all compression approach, which does not account for the varying data volumes across workers. Assigning varying compression ratios to workers with distinct data distributions and volumes is therefore a promising solution. This work derives the convergence rate of distributed SGD with non-uniform compression, which reveals the intricate relationship between model convergence and the compression ratios applied to individual workers. Accordingly, we frame the relative compression ratio assignment as an $n$-variable chi-squared nonlinear optimization problem, constrained by a limited communication budget. We propose DAGC-R, which assigns conservative compression to workers handling larger data volumes. Recognizing the computational limitations of mobile devices, we propose the DAGC-A, which is computationally less demanding and enhances the robustness of compression in non-IID scenarios. Our experiments confirm that the DAGC-R and DAGC-A can speed up the training speed by up to $25.43\%$ and $16.65\%$  compared to the uniform compression respectively, when dealing with highly imbalanced data volume distribution and restricted communication.

\end{abstract}

\begin{IEEEkeywords}
Federated Learning, Non-IID, Data-Aware Gradient Compression
\end{IEEEkeywords}

\section{Introduction} \label{introduction}

\IEEEPARstart{W}{ith} the widespread use of mobile devices and the progress in machine learning \cite{gpt4-exp}, there is a burgeoning interest in distributed machine learning (DML) using these portable platforms. FL is an increasingly important DML framework that addresses the critical need for data privacy in model training across multiple mobile devices. Despite its potential, this paradigm faces significant challenges due to communication bottlenecks, particularly as the number of devices scales up. Gradient compression has been identified as an effective solution to this challenge, by reducing the communication volume and offering a cost-effective option in bandwidth-limited and per-traffic billing mobile environments.

However, in non-IID scenarios, the performance of lossy gradient compression algorithms worsens, leading to poorer model convergence when contrasted with IID datasets \cite{fedavg,niid2020icml}. For instance, the same gradient compression algorithm \cite{hsieh2017gaia} (with the hyperparameter set to $10\%$) reduces the accuracy by only $0.7\%$ compared to the bulk synchronous parallel (BSP) \cite{bsp} as a baseline in IID scenarios. However, this gap widens significantly to a $10.4\%$ decrease in accuracy under non-IID conditions \cite{niid2020icml}. The reason for the drop in accuracy is that it uses the same aggressive  gradient compression ratios for different workers, ignoring the fact that different workers usually have different data volumes and distributions \cite{fed_learning,fednova,li2020fedopt}.  
In real-world non-IID scenarios, mobile devices are geographically distributed, and each worker collects its own dataset, resulting in skewed data distributions and volumes \cite{niidbench,niid2020icml}. To illustrate, within the Flickr-mammal dataset (denoted as Flickr in the following) \cite{niid2020icml}, the worker with the largest data volume has $78\%$ more samples than the worker with the second largest data volume (divided by subcontinent). Similarly, in the Google Landmark dataset v2 \cite{Googlelandmark}, the difference is even greater. The worker with the most data has at least $213\%$ more images than its peers, with the division based on continent.

To facilitate our discussion, we introduce the terms \textit{large workers} and \textit{small workers} to refer to workers with large and small amounts of data, respectively. We use \textit{worker size} to denote the quantity of local data samples. Current gradient compression algorithms often neglect the variation in \textit{worker size}. Even in studies like SkewScout \cite{niid2020icml}, which suggests adaptive methods to adjust gradient compression ratios based on data distribution differences, \textit{large workers} are subjected to the same stringent compression strategy as their smaller counterparts. This leads to the loss of vital information that could otherwise help the model converge faster.

Our empirical analysis reveals important findings for developing a data-volume-aware gradient compression method. Primarily, we find that a \textit{one-size-fits-all} compression strategy falls short in non-IID settings with communication constraints, as workers with diverse data volumes and distributions require tailored compression ratios. Furthermore, a compression strategy that assigns higher compression ratios\footnote{In this work, the compression ratio is defined as the ratio of the compressed data divided by the uncompressed data, referring to the gradient compression part of Sec. II in \cite{dc2}.} to \textit{large workers} can reduce the number of training iterations needed to achieve the same accuracy, in contrast to a uniform strategy. Leveraging these findings, we advocate for an adaptive algorithm that adjusts compression ratios based on the \textit{worker size}, thereby optimizing the balance between compression efficiency and communication overhead. \IEEEpubidadjcol

The relative compressor is one of the most popular types of gradient sparse compressors, known for achieving efficient compression compared to other compressors like quantization \cite{qsgd} or low-rank \cite{vogels2019powersgd}. In relative compressors, we can directly determine the compression ratio, and the compressor will transmit the corresponding number of elements. The main technical challenge in designing the adaptive relative compressor is: \textit{given a fixed total communication budget, how to determine the compression ratio for each worker?} This budget constraint implies that, during each iteration, the total traffic sent from all clients to the server must not exceed a fixed limitation\footnote{We focus solely on compressing the communication from clients to the server, also known as upstream communication. This is because communication bottlenecks in the PS architecture typically occur in the upstream direction \cite{dgc,cui2021slashing}. The downstream communication optimization is beyond the scope of this work.}. To tackle this issue, firstly we derive the convergence rate of distributed SGD with error feedback and different gradient compression ratios (denoted as non-uniform D-EF-SGD) with the relative compressors like Top-$k$ \cite{dgc,stich2018sparsified}. Our results demonstrate that in communication-constrained non-IID scenarios, there is a key term that directly affects the convergence rate of non-uniform D-EF-SGD with the relative compressors. Minimizing this term not only speeds up its convergence rate but also makes the algorithm robust to non-IID scenarios. To find a proper assignment of relative compression ratios, we formulate an $n$-variable chi-square nonlinear optimization problem with one constraint, which can be solved by the Lagrange multiplier method and finding the minimum of a one-dimensional function. 

While the process of deriving the optimal compression ratios introduces negligible additional computational cost, the relative compressor has high computational overhead, making it less suitable for resource-constrained mobile environments. In contrast, the absolute compressor \cite{hardthreshold}, which transmits elements with absolute values higher than a fixed threshold, does not allow for precise control over the compression ratio, and it is more efficient than the SOTA relative compressor Top-$k$. The compression cost of Top-$k$ can be up to hundreds of times greater than that of the absolute compressor\footnote{According to Fig.~15(d) in the appendix of the work \cite{sidco}, the compression time of Top-$k$ is hundreds of times that of SIDCo, and the absolute compressor is more effective than SIDCo.} for two main reasons: (1) the absolute compressor has a lower computational complexity of $\mathcal{O}(d)$ compared to the  $\mathcal{O}(d\log k)$ complexity of Top-$k$ selection, where $d$ is the number of the model parameters; (2) Top-$k$ selection performs poorly on accelerators such as GPUs. Therefore, we solve the technical challenge under the absolute compressor similarly. We derive the convergence rate under the absolute compressor and reveal the key factor in the rate. We then formulate the task identifying the optimal threshold as a symmetric optimization problem, solvable using the Lagrange multiplier method. By minimizing the key factor, we can obtain a faster convergence rate without introducing additional hyperparameters.

\begin{figure}[t]
    \centering
    \includegraphics[width=0.98\linewidth,scale=1.0]{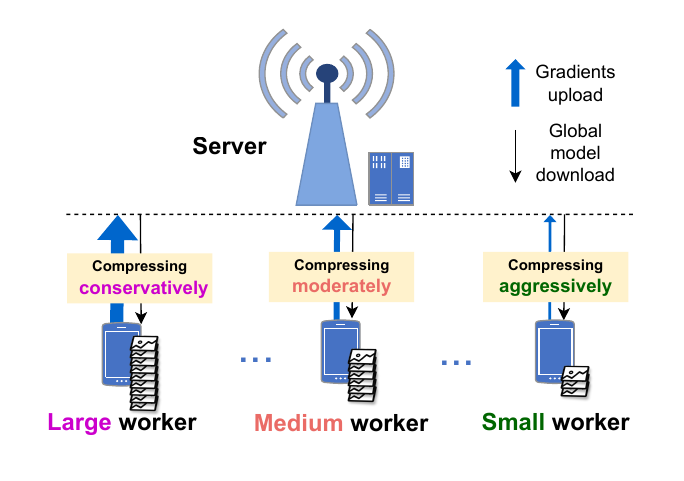}
    \caption{High-level design of DAGC. DAGC sets different compression ratios to workers depending on the \textit{worker size}. \textit{Large workers} (\textit{i.e.}, the workers with large data volumes and similarly to \textit{small} and \textit{medium workers}) are assigned conservative compression ratios, and \textit{small workers} adopt aggressive compression ratios. }
    \label{fig:DAGCOverview}
\end{figure}

Based on these analyses, we propose DAGC (illustrated in Fig.~\ref{fig:DAGCOverview}), a low-cost data-aware adaptive gradient compression algorithm that strategically allocates different compression ratios depending on the \textit{worker size}. DAGC is designed for the realistic non-IID scenario, where the local datasets in the mobile devices are collected based on the location and there is a significant difference in the \textit{worker size}. 
DAGC is composed of DAGC-R for relative compressors and DAGC-A for absolute compressors.  We define the number of workers as $n$, the compression ratio of the relative compressor, the threshold of the absolute compressor, and the training weight of the $i$-th worker as $\delta_i$, $\lambda_i$ and $p_i$ respectively. We have $\frac{\delta_i}{\delta_j}\approx(\frac{p_i}{p_j})^{2/3}$ in DAGC-R and $\frac{\lambda_i}{\lambda_j}=(\frac{p_j}{p_i})^{2/3}$ in DAGC-A. Both align with the conclusion from the measurements, indicating that higher compression ratios should be given to workers with larger $p_i$. The time complexity of DAGC-R (as well as DAGC-A) to find the optimal $\delta_i$ ($\lambda_i$) is $\mathcal{O}(n)$ ($\mathcal{O}(1)$).

In a word, the contributions of our work are summarized as
below:

\begin{itemize}
    \item We experimentally reveal that setting higher compression ratios for \textit{large workers} converges faster than the uniform compression under the fixed and limited communication volume.

    \item We generalize the convergence analysis of D-EF-SGD with both relative \cite{aji2017sparse} and absolute compressors \cite{hardthreshold} to the context of non-uniform compression, where nodes are endowed with different compression ratios and training weights. Under communication-constrained non-IID scenarios, we show the key factor $\frac{\sum_{i=1}^n\frac{p_i}{\sqrt{\delta_i}}}{\sqrt{\delta_{min}}}$ for the relative compressor and $\sum_{i=1}^n p_i^2 \lambda_i^2 $ for the absolute compressor.

    \item We propose two novel adaptive compression algorithms, DAGC-R and DAGC-A, designed for optimizing compression rate allocation in relative and absolute compressors, respectively. DAGC-R is developed by solving an $n$-variable chi-square nonlinear asymmetric optimization problem with a communication constraint. In DAGC-R, it has  $\frac{\delta_i}{\delta_j}\approx(\frac{p_i}{p_j})^{2/3}$. Similarly, DAGC-A is developed by solving an $n$-variable chi-square nonlinear symmetrical optimization problem subject to the same constraint and has $\frac{\lambda_i}{\lambda_j}=(\frac{p_j}{p_i})^{2/3}$. DAGC-R and DAGC-A converge the fastest in communication-constrained non-IID scenarios.

    \item We employ DAGC-R and DAGC-A in both the real-world non-IID and artificially partitioned non-IID datasets. The experimental results confirm the correctness of our theory and show that our design can save up to  $25.43\%$ of iterations to converge to the same accuracy compared to the uniform compression.
    
\end{itemize}
\section{Preliminaries} \label{II}

We specifically concentrate on FL using the D-EF-SGD algorithms alongside sparsification compressors. We aim to overcome the challenges posed by non-IID scenarios with constraint communication in FL. To provide a comprehensive understanding, we will briefly delve into the optimization problem of FL, communication-constraint non-IID scenarios, and gradient compression, including the properties of relative and absolute compressors.

\noindent \textbf{The optimization problem of FL:}
The distribution problem is taken into consideration in this project.
\begin{equation}
f^{*}:= \underset{\textbf{x}}{\min}\left[ f(\textbf{x}):=\sum_{i=1}^np_if_i(\textbf{x})\right] , \nonumber
\end{equation}
In this research, we consider the objective function $f$, which is divided into $n$ terms $f_i, i \in [n]$. $p_i$ is the training weight of the worker $i$. Each $p_i$ value should be greater than or equal to $0$, and the summation of all $p_i$ values equals $1$. To facilitate explanation, we assign sequence numbers to the workers based on their training weights\footnote{The order of the training weight remains the same in the following paper.}, with $p_i \geq p_{i+1}, \forall i \in [n-1] $.

\noindent \textbf{Communication-constraint non-IID scenarios:} In communication-constrained non-IID scenarios, on the data side, data on each node is isolated from others due to privacy and data security. This results in different data volumes and the classification of datasets stored by distributed workers. On the communication side, the bandwidth between mobile devices and the server is limited, often requiring communication across WAN, which incurs high costs. Therefore, achieving efficient training under these constraints can not only speed up the training process but also significantly reduce the expensive WAN communication costs. In this work, communication-constrained cases mean that the average compression ratio is less than or equal to $0.1\%$, and having the same communication budget means that the sum of the communication volume transmitted from all workers to the server per iteration is the same across different compression strategies.

\noindent \textbf{Gradient compression:} Unlike compressing the model \cite{tang2022mixed,tang2023elasticvit} for inference speedup, gradient compression focuses on compressing the gradient 
to reduce the traffic volume during the communication phase.
Concerning compression tactics \cite{grace}, compression approach can be categorized into (i) quantization  \cite {bernstein2018signsgd,wu2018error,qsparse}, which converts high precision to low precision, consequently diminishing the number of transmitted bits; (ii) sparsification \cite{alistarh2018convergence,hardthreshold,alistarh2018convergence,hardthreshold}, which retains solely certain elements of the gradient while assigning $0$ to others; (iii) low-rank \cite{vogels2019powersgd}, which breaks down the gradient matrix to acquire multiple low-rank matrices.

According to whether the compression error (i.e., the norm of the difference between the parameter before and after compression) is independent of the compressed parameter $\textbf{x}$, the compressors can be categorized as absolute or relative compressors \cite{hardthreshold}:

\noindent$\bullet$  \textit{The relative compressor:} A relative compressor is one whose compression error is dependent on the compressed parameter. Classic relative compressors include Top-$k$ and Random-$k$ \cite{dc2,grace}. The input parameter of a relative compressor is the compression ratio $\delta$. The larger $\delta$ is, the more conservative the compression is. We represent the relative compressor as $\textbf{C}_{\delta}$. By definition, $\textbf{C}_{\delta}$ is a mapping that has the property $\mathbb{R}^d \rightarrow \mathbb{R}^d$:
\begin{equation}
    \mathbb{E}_{\textbf{C}_{\delta}}\lVert \textbf{C}_{\delta}(\textbf{x}) - \textbf{x} \rVert^2 \leq (1-\delta) \lVert \textbf{x} \rVert^2. \nonumber
\end{equation}

\noindent$\bullet$  \textit{The absolute compressor:} The compressor error of the absolute compressor is independent of $\textbf{x}$. The representative compressor is the hard-threshold algorithm \cite{hardthreshold}. The input parameter of the absolute compressor is the threshold $\lambda$. The higher $
\lambda$ indicates more aggressive compression. We denote the absolute compressor as $\textbf{C}_{\lambda}$ and definitionally, $\textbf{C}_{\lambda}$ is a mapping: $\mathbb{R}^d \rightarrow \mathbb{R}^d$, having the property:
\begin{equation}\label{absolute property}
    \mathbb{E}_{\textbf{C}_{\lambda}}\lVert \textbf{C}_{\lambda}(\textbf{x}) - \textbf{x} \rVert^2 \leq d \lambda^2, \nonumber
\end{equation}
where $d$ is the number of parameters of the model.

\noindent \textbf{Distributed SGD with error-feedback and sparsification compressors (D-EF-SGD):} D-EF-SGD \cite{stc}  and Distributed Quantized SGD (D-QSGD) \cite{cui2021slashing,QuantityGlobal} are two of the most dominant compressors in FL. In communication-constrained non-IID scenarios, D-EF-SGD has two advantages: First, it achieves lower compression ratios\footnote{$\frac{1}{32}$ is the tiniest compression ratio of D-QSGD  \cite{signsgd,qsgd}, nearly $3.4\%$, while less than $0.1\%$ denotes the compression ratio of sparsification constrain \cite{aji2017sparse}. Meanwhile, the compression ratio of D-QSGD is discrete. It emphasizes to adaptively adjust more complicated ratios. This issue is also discussed in the design of DC2 \cite{dc2}.} (\textit{i.e.}, transmits less information), which is favorable for resource-constrained scenarios; Second, D-EF-SGD has a lower dependence on the heterogeneity of the data, and thus is more suitable for highly skewed non-IID scenarios  \cite{stich2020communication}. For these reasons, we focus on D-EF-SGD rather than D-QSGD.

\section{Motivating Examples} \label{III}

In this section, we aim to validate the following two points through a series of motivating experiments: 1) In communication-constrained environments, a compression strategy with different compression ratios can achieve faster convergence compared to uniform compression. 2) A strategy that sets higher compression ratios for \textit{large workers} usually achieves faster convergence than those for \textit{small workers}, reducing the number of iterations by up to $69.70\%$. To verify this conclusion, two non-uniform compression strategies are used in this paper: 1) \textit{Non-uniform compression I give higher compression ratios to large workers} and lower compression ratios to small workers; 2) \textit{Non-uniform compression II gives lower compression ratios to large workers} and higher compression ratios to small workers.

\begin{figure}[t]
\centering
\subfloat[Logistic@FMNIST]{
\includegraphics[width=0.45\linewidth]{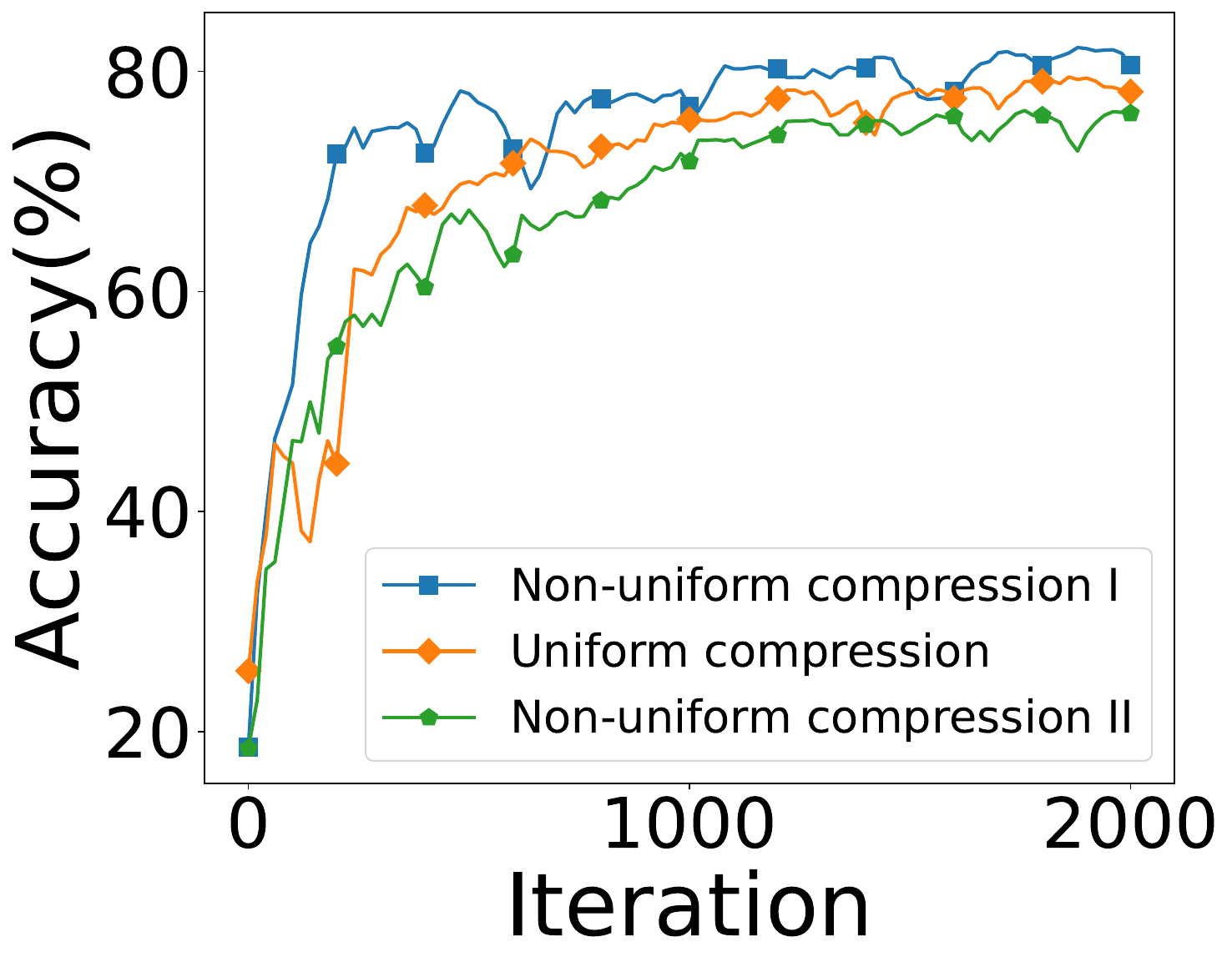}\label{fig:2:a}}
\hspace{0.01\linewidth}
\subfloat[LSTM@SCs]{\includegraphics[width=0.45\linewidth]{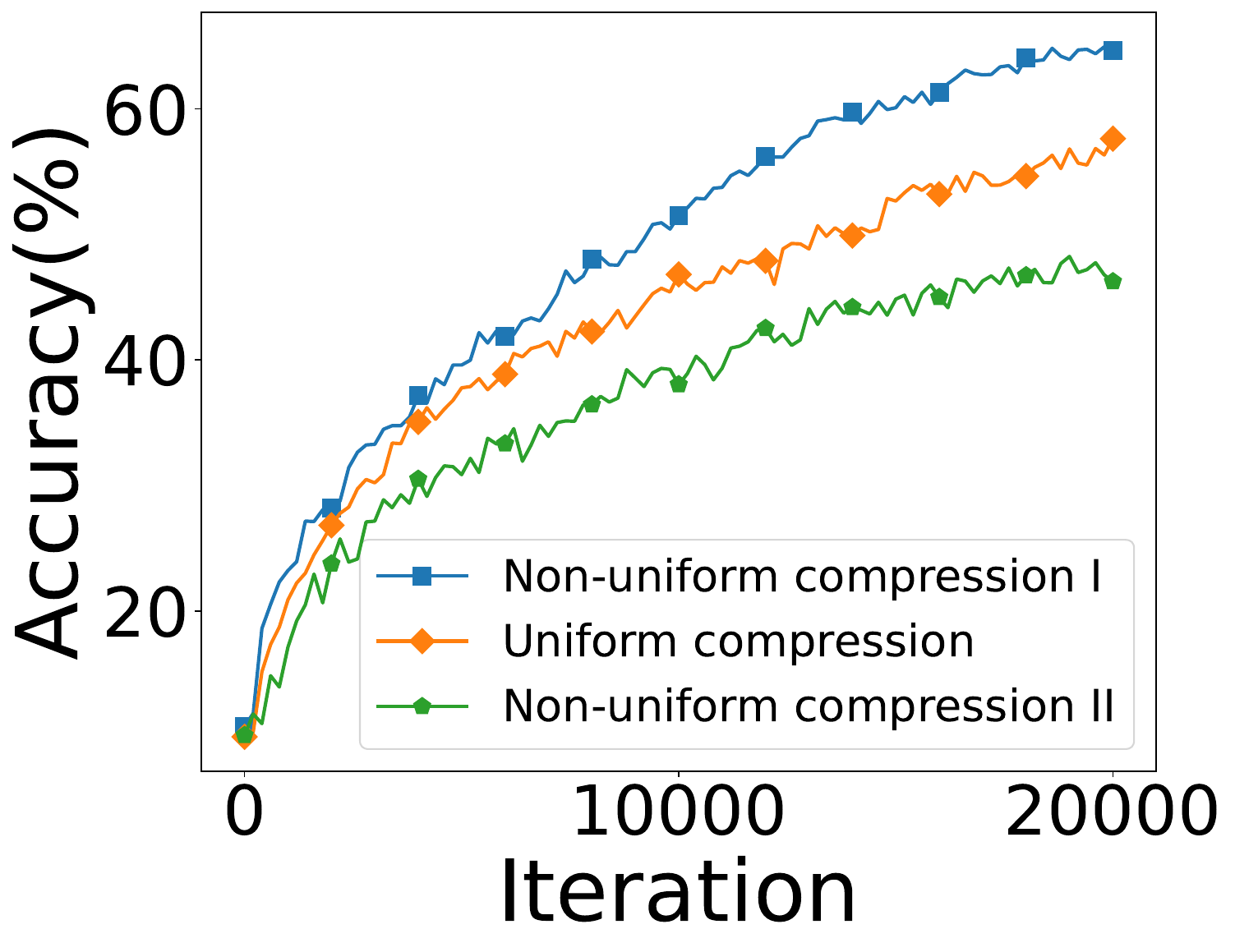}\label{fig:2:b}}
\caption{The accuracy curves (Accuracy vs. Iterations) of  Logistic@FMNIST (a) and LSTM@SCs (b) using different relative compression strategies. In scheme I (as well as scheme II), large workers are set lower (higher) compression ratios. The uniform compression is a one-size-fits-all strategy. Among these three strategies, non-uniform compression scheme I exhibits optimal performance.}
\label{fig:2}
\end{figure}

To provide empirical evidence for our research, we performed two tasks: Fashion-MNIST \cite{fmnist} (denoted as FMNIST) using Logistic and Speech Commands \cite{warden2018speech} (denoted as SCs) using LSTM. This ensures that our experimental findings are generalizable. To gain a better understanding of the results, we divide the workers into two categories: \textit{large workers} and \textit{small workers}. Our setup consists of $11$ workers, with one \textit{large worker} with datasets accounting for $p_{large}$ of the global dataset, and $10$ \textit{small workers}, each with datasets accounting for $p_{small} = \frac{1-p_{large}}{10}$, $p_{large} = 50\%$.
In these experiments, we impose communication constraints, specifically an average compression ratio of the aggressive ratio $\delta_{min}=0.1\%$ as described in the appendix experiments \cite{hardthreshold}, while maintaining consistent total communication volume (i.e., fixing $\sum_{i=1}^n \delta_i $). To simulate the non-IID scenario, the data points are generated based on the Dirichlet distribution, with the parameter set to $0.5$. 
 
\noindent \textbf{Logistic on FMNIST:} 
It can be observed from Fig.~\ref{fig:2:a} that the initial stages witness faster progress in non-uniform compression I ($\delta_1=1\%$, and $\delta_i=0.01\%$, $i\in [2,n]$) when compared to two alternative strategies. In order to achieve accuracies of $50\%$, $60\%$, $70\%$, and $80\%$, the iteration reduction for scheme I are $54.55\%$, $41.67\%$, $62.96\%$, and $62.50\%$, respectively, compared to the uniform compression ($\delta_i=0.1\%$, $i \in [1,n]$). It is noteworthy that under such circumstances, the performance of scheme I is optimized, whereas non-uniform compression II ($\delta_1=0.01\%$, and $\delta_i=0.11\%$, $i\in [2,n]$) experiences a relatively weaker performance. 

\noindent \textbf{LSTM on SCs:} Fig.~\ref{fig:2:b} demonstrates that, in the context of fixed communication volume, the model utilizing the non-uniform compression strategy I exhibits the fastest convergence. To attain accuracy levels of $50\%$, $55\%$, $60\%$, and $65\%$, the non-uniform compression strategy I diminishes the training time by $7.95\%$, $19.17\%$, $19.75\%$, and $27.06\%$, respectively, as compared to the uniform strategy with $p_{large}=50\%$. The convergence rate of scheme II is significantly worse than the other two strategies. 

In conclusion, the aforementioned empirical findings affirm the inadequacy of uniform compression as an optimal strategy when confronted with discrepancies in worker size. Conversely, the proposed approach, which assigns higher compression ratios to \textit{large workers}, stands as an efficacious means of enhancing training speed.

\section{Theoretical Analysis within the Relative Compressor} \label{DAGC-R}
In this section, our primary focus is to address the central issue of this research: \textbf{Given the total amount of communication, how can we theoretically ascertain the best compression ratio for each worker using the relative compressor?}

\begin{table}[!t]
\centering
\caption{Notation list.}
\label{table:notation}
\scalebox{1.2}{
\begin{tabular}{ |c|c| } 
\hline
Notation & Description \\
\hline
$n$ &  the number of workers\\
 \hline
$p_i$ & the training weight of the $i$-th node\\
\hline
\multirow{3}{*}{$\delta_i / \lambda_i $} & the compression ratio/threshold  \\& of node $i$ under the \\& relative/absolute compressor\\
\hline
\multirow{2}{*}{$\textbf{C}_{\delta_i} / \textbf{C}_{\lambda_i}$} & the relative/absolute compressor \\& and the hyper-parameter is $\delta_i / \lambda_i$\\
\hline
 $\textbf{x}_t$ &  the global model in the $t$-th iteration \\
 \hline
  $\gamma$ &  learning rate \\
 \hline
 
 $\textbf{g}^i(\textbf{x}_t)$ & the stochastic gradient of $\textbf{x}_t$ \\
\hline
\multirow{2}{*}{$\textbf{e}_t^{i}$} & the local error term of node $i$  \\& in the $t$-th iteration \\
\hline
\multirow{3}{*}{ $\hat{\Delta}_t^i$} & the compressed gradients \\& transferred    to the server from \\& node $i$  in  the  $t$-th iteration \\
 \hline
 \multirow{3}{*}{ $\hat{\Delta}_t^i$} & the compressed gradients \\& transferred    to the server from \\& node $i$  in  the  $t$-th iteration \\
 \hline
  \multirow{3}{*}{$\zeta_i$} & the distance of node $i$  \\& from the local distribution \\& to the global distribution \\
 \hline
  \multirow{2}{*}{$\sigma^2$} & the upper variance bound \\& of the  noise of gradients \\
\hline
 
\end{tabular}
}
\end{table}

Initially, we outline the convergence speed of non-uniform D-EF-SGD with the relative compressor, taking into account the non-convex, convex, and strongly convex scenarios. (Theorems in Sec.~\ref{IV-B} and the proof in Appendix)

Subsequently, we represent this challenge as an $n$-variable chi-square nonlinear asymmetrical optimization problem that comes with one constraint. (\textbf{Theorem~\ref{theorem:4}} in Sec.~\ref{IV-D}) 

Lastly, we propose our design DAGC-R in Sec.~\ref{IV-E}.

We tabulate the notations in Table~\ref{table:notation}. Additionally, we showcase the pseudo-code of non-uniform D-EF-SGD in Algorithm~\ref{algo:1}.

\begin{algorithm}[t]
    \SetAlgoLined
	\caption{Non-uniform D-EF-SGD}
	\label{algo:1}
    \KwIn{$n$, $\gamma$, $p_i$, compressor $\textbf{C}$, compression parameters $\delta_1,\ldots,\delta_n$ or $\lambda_1,\ldots,\lambda_n$, 
    initial parameters $\textbf{x}_0$, 
      initial local error $\textbf{e}_0^i = \textbf{0}_d$ } 
	\KwOut{$\textbf{x}_T$}
	\For{$t = 0, \ldots , T - 1 $}{
	    \tcc{Worker does}
        \For{$i = 1, \ldots, n $}{
            Receive $\textbf{x}_t$ from the server\;
    	    $\textbf{g}_t^i := \textbf{g}^i(\textbf{x}_t)$\;
            \eIf{$\textbf{C}$ is the relative compressor}{
                 $\hat{\Delta}_t^i := \textbf{C}_{\delta_i} (\textbf{e}_t^i + \textbf{g}_t^i ) $\;
                 }{
                 $\hat{\Delta}_t^i := \textbf{C}_{\lambda_i} (\textbf{e}_t^i + \textbf{g}_t^i ) $\;
                 }
    	    $\textbf{e}_{t+1}^i := \textbf{e}_t^i + \textbf{g}_t^i - \hat{\Delta}_t^i $\;
    	    Upload $\hat{\Delta}_t^i$\;
    	    }
    	    {
    	    \tcc{Server does}
    	    Gather all $\hat{\Delta}_t^i$\;
    	    $\textbf{x}_{t+1} := \textbf{x}_t - \gamma\sum_{i=1}^n p_i\hat{\Delta}_t^i$\;
    	    Send $\textbf{x}_{t+1}$ to all workers
    	    }
	}
	\textbf{Return} $\textbf{x}_T$\;
\end{algorithm}

\subsection{Assumptions}\label{IV-A}

We assume functions are $L$-smooth with the gradient noise of SGD presumed to exhibit zero mean and a variance of $\sigma^2$. We gauge data heterogeneity using constants $\zeta_i^2 > 0$ and $Z^2 \geq 1$, which cap the variance among the $n$ workers. In \textbf{Theorem~\ref{theorem:2},~\ref{theorem:3}}, we posit that objective functions are $\mu$-strongly convex. Detailed assumptions are shown below.

\noindent \textbf{Assumption 1} ($L$-smoothness). We assume $L$-smoothness of $f_i, i\in [n]$, that is, for all $\textbf{x},\textbf{y} \in \mathbb{R}^d$:
\begin{equation}
\lVert \nabla f(\textbf{y}) - \nabla f(\textbf{x}) \rVert \leq L\lVert \textbf{y} - \textbf{x} \rVert.
\end{equation}

\noindent \textbf{Assumption 2} ($\mu$-strongly convexity). We assume $\mu$-strong convexity of $f_i, i\in [n]$, that is, for all  $\textbf{x},\textbf{y} \in \mathbb{R}^d$:
\begin{equation}
f(\textbf{x}) - (\textbf{y}) \geq  \langle \nabla f(\textbf{y}),\textbf{x}-\textbf{y} \rangle + \frac{\mu}{2} \lVert \nabla f(\textbf{y}) - \nabla f(\textbf{x}) \rVert^2.
\end{equation}

\noindent \textbf{Assumption 3} (Bounded gradient noise). We assume that we have access to stochastic gradient oracles $g^i(\textbf{x}): \mathbb{R}^d \rightarrow  \mathbb{R}^d$ for each $f_i,i\in [n]$. For simplicity we only consider the instructive case of uniformly bounded noise for all $\textbf{x} \in \mathbb{R}^d, i \in [n]$:
\begin{equation}
g^i(\textbf{x})=\nabla f_i(\textbf{x})+\boldsymbol{\xi} ^i,
\quad \mathbb{E}_{\boldsymbol{\xi} ^i}\boldsymbol{\xi} ^i = \textbf{0}_d, \quad \mathbb{E}_{\boldsymbol{\xi} ^ i} \lVert \boldsymbol{\xi} ^i \rVert^2 \leq \sigma^2.
\end{equation}

 \noindent \textbf{Assumption 4} (Measurement of data heterogeneity). We measure data dissimilarity by constants $\zeta_i^2 \geq 0,Z^2 \geq 1$ that bound the variance across the $n$ nodes. We have:
\begin{equation}
\lVert \nabla f_i(\textbf{x}) \rVert^2 \leq \zeta_i^2+Z^2\lVert\nabla f(\textbf{x})\rVert^2, \quad \forall \textbf{x} \in \mathbb{R}^d,i \in [n]. \nonumber
\end{equation}

This is similar to the assumption in previous work \cite{cui2022optimal, stich2020communication}.

\subsection{Convergence rate of non-uniform D-EF-SGD with the relative compressor}\label{IV-B}

\begin{theorem}[Non-convex convergence rate of non-uniform D-EF-SGD with the relative compressor]
\label{theorem:1}
Consider a function $f$, which maps from $\mathbb{R}^d$ to  $\mathbb{R}$, and is $L$-consistent. We can find a learning rate $\gamma$ such that  $\gamma \leq \frac{1}{4LZ} \frac{\delta_{min}}{\sqrt{n C_Z}}$, where  $C_Z = \sum_{i=1}^n\frac{\delta_{min}}{\delta_i}p_i^2$. This means that the number of 

\begin{equation}
\resizebox{.95\hsize}{!}{
$
\begin{aligned}
    \mathcal{O}(&\frac{\sigma^2\sum_{i=1}^n p_i^2}{\epsilon^2}+\frac{\sqrt{n}(\zeta\sum_{i=1}^n\frac{p_i}{\sqrt{\delta_i}}+\sigma\sqrt{\sum_{i=1}^n p_i^2})}{\epsilon^{3/2} \sqrt{\delta_{min}} }\\
    &+\frac{\sqrt{n} Z\sum_{i=1}^n\frac{p_i}{\sqrt{\delta_i}}}{\epsilon \sqrt{\delta_{min}}})\cdot L F_0 
\end{aligned}
$
}
\end{equation}
iterations of non-uniform D-EF-SGD with the relative compressor ensures $\mathbb{E}f(\textbf{x}_{final})-f^* \leq \epsilon$, where $F_0$ is at least $ f(\textbf{x}_0)-f^*$, and $\textbf{x}_{final} = \textbf{x}_t$ refers to a version $\textbf{x}_t$ from the set $\left\{\textbf{x}_0, \ldots, \textbf{x}_{T-1}\right\}$, picked randomly.

\end{theorem}

\begin{theorem} [Convex convergence rate of non-uniform D-EF-SGD with the relative compressor, \textit{i.e.}, $\mu = 0$]
\label{theorem:2}
Consider a function $f$, which maps from $\mathbb{R}^d$ to  $\mathbb{R}$, and is $L$-consistent and $\mu$-convex. We can find a learning rate $\gamma$ such that  $\gamma \leq \frac{1}{14LZ} \frac{\delta_{min}}{\sqrt{n C_Z}}$, where  $C_Z = \sum_{i=1}^n\frac{\delta_{min}}{\delta_i}p_i^2$. This means that the number of 

\begin{equation}
\resizebox{.9\hsize}{!}{
$
\begin{aligned}
    \mathcal{O}(&\frac{\sigma^2\sum_{i=1}^n p_i^2}{\epsilon^2}+\frac{\sqrt{n L}(\zeta\sum_{i=1}^n\frac{p_i}{\sqrt{\delta_i}}+\sigma\sqrt{\sum_{i=1}^n p_i^2})}{\epsilon^{3/2}\sqrt{\delta_{min}} }\\
    &+\frac{\sqrt{n} L Z\sum_{i=1}^n\frac{p_i}{\sqrt{\delta_i}}}{\epsilon\sqrt{\delta_{min}}})\cdot R_0^2 
\end{aligned}
$
}
\end{equation}
iterations of non-uniform D-EF-SGD with the relative compressor ensures $\mathbb{E}f(\textbf{x}_{final})-f^* \leq \epsilon$, where $R_0^2$ is at least $ \lVert \textbf{x}_0 -\textbf{x}_* \rVert$, and $\textbf{x}_{final} = \textbf{x}_t$ refers to a version $\textbf{x}_t$ from the set $\left\{\textbf{x}_0, \ldots, \textbf{x}_{T-1}\right\}$, picked randomly.

\end{theorem}

\begin{theorem} [Strong convex convergence rate of non-uniform D-EF-SGD with the relative compressor, \textit{i.e.}, $\mu > 0$]
\label{theorem:3}
Consider a function $f$, which maps from $\mathbb{R}^d$ to  $\mathbb{R}$ , and is $L$-consistent and $\mu$-convex. We can find a learning rate $\gamma$ such that  $\gamma \leq \frac{1}{14LZ} \frac{\delta_{min}}{\sqrt{n C_Z}}$, where  $C_Z = \sum_{i=1}^n\frac{\delta_{min}}{\delta_i}p_i^2$. This means that the number of

\begin{equation}
\resizebox{.9\hsize}{!}{
$
\begin{aligned}
    \tilde{\mathcal{O}}(&\frac{\sigma^2\sum_{i=1}^n p_i^2}{\mu \epsilon}+\frac{\sqrt{n L}(\zeta\sum_{i=1}^n\frac{p_i}{\sqrt{\delta_i}}+\sigma\sqrt{\sum_{i=1}^n p_i^2})}{\mu \sqrt{\delta_{min} \epsilon}}\\
    &+\frac{\sqrt{n} L Z\sum_{i=1}^n\frac{p_i}{\sqrt{\delta_i}}}{\mu \sqrt{\delta_{min}}})
\end{aligned}
$
}
\end{equation}
iterations of non-uniform D-EF-SGD with the relative compressor ensures $\mathbb{E}f(\textbf{x}_{final})-f^* \leq \epsilon$, and $\textbf{x}_{final} = \textbf{x}_t$ refers to a version $\textbf{x}_t$ from the sequence $\left\{\textbf{x}_0, \ldots, \textbf{x}_{T-1}\right\}$, selected probabilistically based on $(1-\min\left\{\frac{\mu \gamma}{2},\frac{\delta_{min}}{4}\right\})^{-t}$.

\end{theorem}

\subsection{Analysis of the convergence rate in communication-constrained non-IID scenarios}\label{IV-C}

To facilitate, we denote the order of magnitude of $a$ as $OM(a)$, i.e., $OM(a)= \lfloor  \log_{10} a \rfloor $. In this way, we have the following inequalities:
\begin{equation}
OM(\epsilon) \geq -4 ,
\label{OM-1}
\end{equation}
\begin{equation}
OM(n) \geq 1 ,
\label{OM-4}
\end{equation}
\begin{equation}
OM(\sigma) \leq OM(\zeta) +1 ,
\label{OM-2}
\end{equation}
\begin{equation}
OM(\delta) \leq -3.
\label{OM-3}
\end{equation}

Eq.~\ref{OM-1} and Eq.~\ref{OM-4} are the default experimental settings. The Eq.~\ref{OM-1} means that the model converges when $\epsilon = 10^{-4}$ according to the previous work \cite{koloskova2022sharper}. $n$ is typically taken from $10$ to $1,000$, so we get the Eq.~\ref{OM-4}.

Eq.~\ref{OM-2} and Eq.~\ref{OM-3} are based on the communication-constrained non-IID scenario discussed in this paper. The Eq.~\ref{OM-2} means that the bias due to the non-IID datasets is larger than the noise of the gradient,  based on the accuracy degradation of the severe non-IID problem \cite{niid2020icml, niidbench}. In other words, if Eq.~\ref{OM-2} does not hold, the negative effect of the non-IID problem is negligible, which is beyond the scope of this work. Due to $\delta \leq 0.1\%$ in the communication-constrained case, we get Eq.~\ref{OM-3}.

Based on the above inequalities, we have
\begin{equation}
\begin{aligned}
&OM(\zeta\sum_{i=1}^n\frac{p_i}{\sqrt{\delta_i}}) \\ &=  OM(\zeta) + OM(\sum_{i=1}^n p_i) - \frac{1}{2} OM(\delta)\\ &\geq (OM(\sigma)-1) + OM(\sqrt{\sum_{i=1}^n p_i^2}) + OM(\sqrt{n}) + \frac{3}{2} \\ &>OM(\sigma\sqrt{\sum_{i=1}^n p_i^2}), \nonumber
\end{aligned}
\end{equation}
so the convergence in the \textbf{Theorem~\ref{theorem:3}} can be written into 

\begin{equation}
\begin{aligned}
    \tilde{\mathcal{O}}(&\frac{\sigma^2\sum_{i=1}^n p_i^2}{\mu \epsilon}+\frac{\sqrt{n L}(\zeta\sum_{i=1}^n\frac{p_i}{\sqrt{\delta_i}})}{\mu \sqrt{\delta_{min} \epsilon}}\\
    &+\frac{\sqrt{n} L Z\sum_{i=1}^n\frac{p_i}{\sqrt{\delta_i}}}{\mu \sqrt{\delta_{min}}}). \nonumber
\end{aligned}
\end{equation}

For the sake of simplicity, we introduce a function of $n$ variables, $\Phi(\delta_1,\ldots,\delta_n)=\frac{\sum_{i=1}^n\frac{p_i}{\sqrt{\delta_i}}}{\sqrt{\delta_{min}}}$. Then the convergence rate of \textbf{Theorem~\ref{theorem:3}} can be further simplified as 

\begin{equation}
\begin{aligned}
\label{simple-convergence}
    \tilde{\mathcal{O}}(&\frac{\sigma^2}{\epsilon}+\frac{\zeta\Phi}{\sqrt{\epsilon}}+\Phi). \nonumber
\end{aligned}
\end{equation}

Similarly, \textbf{Theorem~\ref{theorem:1}, \ref{theorem:2}} can be  both simplified as 

\begin{equation}
\begin{aligned}
    \mathcal{O}(&\frac{\sigma^2}{\epsilon^2}+\frac{\zeta\Phi}{\epsilon^{3/2}}+\frac{\Phi}{\epsilon}). \nonumber
\end{aligned}
\end{equation}

The difference between \textbf{Theorem~\ref{theorem:1}} and \textbf{Theorem~\ref{theorem:2}} is that there is a coefficient $\sqrt{L}$ on the second term and $L$ on the third term in Theorem~\ref{theorem:2}. It does not matter that they can be represented within the same paradigm. 

We dissect the convergence speed under two circumstances:

\noindent$\bullet$ \textit{Without gradient noise ($\sigma=0$)}: non-uniform D-EF-SGD with the relative compressor shows sub-linear convergence at a pace of  $ \mathcal{O}(\frac{\zeta\Phi}{\sqrt{\epsilon}}+\Phi)$ (as well as $ \mathcal{O}(\frac{\zeta\Phi}{\epsilon^{3/2}}+\frac{\Phi}{\epsilon}$) in the strongly convex cases (non-convex and convex cases). Minimizing $\Phi$ leads to the best convergence.

\noindent$\bullet$ \textit{With gradient noise ($\sigma \neq 0$)}: Notably, the second term $\frac{\zeta\Phi}{\sqrt{\epsilon}}$ cannot be ignored, due to $OM(\frac{1}{\epsilon}) \leq OM(\frac{1}{\sqrt{\epsilon \delta}})$. $\Phi$ is still a key factor in the convergence rate. Reducing $\Phi$ can not only speed up the convergence rate but also mitigate the negative influence of non-IID scenarios. 

Overall,  minimizing $\Phi$ can (1) improve the convergence rate of non-uniform D-EF-SGD with the relative compressor; and (2)  make the algorithm robust to non-IID scenarios, in communication-constrained non-IID scenarios regardless of the convexity.                                        
\subsection{The optimal compression ratios}\label{IV-D}

\begin{theorem} [Optimal $\delta_i$] \label{theorem:4}
Under the premise that the overall communication traffic is fixed, \textit{i.e.}, $\sum_{i=1}^n \delta_i=n \Bar{\delta}$, the following equation emerges. We set $\delta_j=\delta_{min}=\min\{\delta_1,\ldots,\delta_n\}$ and denote $P:=\sum_{i=1}^n p_i^{2/3}$. The minimal value of  $\Phi(\delta_1,\ldots,\delta_n)$ can be split into two distinct situations:

\noindent$\bullet$ \textit{$j$ is not equal to $n$}
\begin{equation}
\Phi(\delta_1,\ldots,\delta_n)  \geq \frac{1}{n\Bar{\delta}}(p_j(1+Q_j)+p_n Q_j(1+Q_j)),
\label{theorem:4-1}
\end{equation}
where $Q_j=\frac{P-p_j^{2/3}}{p_n^{2/3}}$. It takes the equal  sign when $\delta_j=\frac{n\Bar{\delta}}{Q_j+1}$ and $\delta_i=\frac{n\Bar{\delta}}{Q_j+1}\frac{p_i^{2/3}}{p_n^{2/3}}, i \neq j$.

\noindent$\bullet$ \textit{$j$ is equal to $n$}
\begin{equation}
\Phi(\delta_1,\ldots,\delta_n)  \geq \frac{1}{n\Bar{\delta}}(p_j(1+Q_j)+p_{n-1} Q_j(1+Q_j)),
\label{theorem:4-2}
\end{equation}
where $Q_j=\frac{P-p_j^{2/3}}{p_{n-1}^{2/3}}$. It takes the equal sign when $\delta_j=\frac{n\Bar{\delta}}{Q_j+1}$ and $\delta_i=\frac{n\Bar{\delta}}{Q_j+1}\frac{p_i^{2/3}}{p_{n-1}^{2/3}}, i \neq j$.

\end{theorem}

\begin{remark}
This theorem reduces the problem of finding the optimal $\delta_i$, to get the minimal value of $\Phi$, from a continuous space into a discrete sub-space with only $n$ points. Since traversing the continuous space is impractical, this theorem does simplify the problem, making it solvable.
\end{remark}

\begin{remark}
In IID scenarios, where $p_i = p_j$ and $\zeta_i = 0$ for all $i, j \in [n]$, it follows that $Q_i = n - 1$ for all $i \in [n]$, leading to $\delta_i = \bar{\delta}$ for all $i \in [n]$. This implies that the optimal compression strategy in IID scenarios is the uniform compression.
\end{remark}

\subsection{DAGC-R and its implement in FL}\label{IV-E}

\begin{algorithm}[t]
    \SetAlgoLined
	\caption{DAGC-R}
	\label{algo:2}
    \KwIn{$n$, $p_i$, average compression ratio $\Bar{\delta}$} 
	\KwOut{$\delta_1,\ldots,\delta_n$}
	\tcc{The value of $\phi_j$ represents the minimal $\Phi(\delta_1,\ldots,\delta_n)$ when $\delta_j$ is the least among all $\delta_i$.}
	Set $\phi_{min} = +\infty$\; 
	\For{$j = n, n-1, \ldots, 1$}{
	    \eIf{$j == n$}{
	     Calculate $\phi_j$ using the right side of Eq.~\eqref{theorem:4-2}\;
	    }{\eIf{$p_j == p_{j-1}$}{
	    \tcc{When weights are repeated, bypass this computation.}
	    $\phi_j=\phi_{min}$\;
	    }{
	    Calculate $\phi_j$ using the right side of Eq.~\eqref{theorem:4-1}\;
	    }
	    }
	    \If{$\phi_j<\phi_{min}$}{
	    $\phi_{min}=\phi_j$ and adjust the optimal values for $\delta_1,\ldots,\delta_n$\;
	    }
	}
	\textbf{Return} $\delta_1,\ldots,\delta_n$\;
\end{algorithm}

The pseudo-code of DAGC under the relative compressor (denoted as DAGC-R) is presented in Algorithm~\ref{algo:2}. DAGC-R is structured in the following manner:  1) it uses \textbf{Theorem~\ref{theorem:4}} to derive $n$ local optimal solutions from a continuous space, and 2) it gets the global optimal solution by traversing $n$ local optimal solutions. 

This procedure has negligible extra overhead compared to conventional gradient compression algorithms. It involves only one local computation at the server, with a time complexity of $\mathcal{O}(n)$ to derive $n$ local optima, and a single communication step to send the optimal compression ratios to workers, both completed before training. DAGC-R does not require external nodes in a parameter server architecture, the server handles the calculations, while in decentralized training, any node can temporarily act as the server. A larger number of devices does not induce extra cost, so DAGC has good scalability.
\section{Theoretical Analysis within the Absolute Compressor} \label{DAGC-A}

In this section, we address the technical challenge encountered under the absolute compressor, \textit{i.e.}, \textbf{Given the limited communication budget, what is the optimal $\lambda_i$?} 

Initially, we outline the convergence speed of non-uniform D-EF-SGD with absolute compressors (Theorems in Sec.~\ref{V-A}). Then, we formulate the challenge as an $n$-variable chi-square nonlinear symmetrical optimization problem with the traffic budget. We propose DAGC-A based on \textbf{Theorem~\ref{theorem-DAGC-A}}. The details to prove theorems are shown in the Appendix.

\subsection{Convergence rate of non-uniform D-EF-SGD with the absolute compressor} \label{V-A}

\begin{theorem}[Non-convex convergence rate of non-uniform D-EF-SGD with the absolute compressor]
\label{non-convex D-EF-SGD-A}
Consider a function $f$, which maps from $\mathbb{R}^d$ to  $\mathbb{R}$, and is $L$-consistent. We can find a learning rate $\gamma$ such that $\gamma \leq \frac{1}{4L}$. Then there exists a stepsize $\gamma \leq \frac{1}{4L}$. This means that the number of 

\begin{equation}
\resizebox{.9\hsize}{!}{
$
\begin{aligned}
    \mathcal{O}(\frac{\sigma^2\sum_{i=1}^n p_i^2}{\epsilon^2}+\frac{\sqrt{nd\sum_{i=1}^n p_i^2 \lambda_i^2}}{\epsilon^{\frac{3}{2}}}+\frac{1}{\epsilon})\cdot LF_0 
\end{aligned}
$
}
\end{equation}
iterations of non-uniform D-EF-SGD with the absolute compressor ensures $\mathbb{E}f(\textbf{x}_{final})-f^* \leq \epsilon$, where $F_0$ is at least $ f(\textbf{x}_0)-f^*$, and $\textbf{x}_{final} = \textbf{x}_t$ refers to a version $\textbf{x}_t$ from the set $\left\{\textbf{x}_0, \ldots, \textbf{x}_{T-1}\right\}$, picked randomly.
\end{theorem}

\begin{theorem}[Convex convergence rate of non-uniform D-EF-SGD with the absolute compressor, \textit{i.e.}, $\mu = 0$]
\label{convex D-EF-SGD-A}
Consider a function $f$, which is mapping from $\mathbb{R}^d$ to  $\mathbb{R}$, $L$-consistent and $\mu$-convex. We can find a learning rate $\gamma$ such that  $\gamma \leq \frac{1}{4L}$. This means that the number of 

\begin{equation}
\resizebox{.9\hsize}{!}{
$
\begin{aligned}
    \mathcal{O}(\frac{\sigma^2\sum_{i=1}^n p_i^2}{\epsilon^2}+\frac{\sqrt{nLd\sum_{i=1}^n p_i^2 \lambda_i^2}}{\epsilon^{\frac{3}{2}}}+\frac{L}{\epsilon})\cdot R_0^2 
\end{aligned}
$
}
\end{equation}
iterations of non-uniform D-EF-SGD with the absolute compressor ensures $\mathbb{E}f(\textbf{x}_{final})-f^* \leq \epsilon$, where $R_0^2$ is at least $ \lVert \textbf{x}_0 -\textbf{x}_* \rVert$, and $\textbf{x}_{final} = \textbf{x}_t$ refers to a version $\textbf{x}_t$ from the set $\left\{\textbf{x}_0, \ldots, \textbf{x}_{T-1}\right\}$, picked randomly.

\end{theorem}

\begin{theorem}[Strong convex convergence rate of non-uniform D-EF-SGD with the absolute compressor, \textit{i.e.}, $\mu > 0$]
\label{strong convex D-EF-SGD-A}
Consider a function $f$, which is mapping from $\mathbb{R}^d$ to  $\mathbb{R}$, $L$-consistent and $\mu$-convex. We can find a learning rate $\gamma$ such that  $\gamma \leq \frac{1}{4L}$. This means that the number of 

\begin{equation}
\resizebox{.9\hsize}{!}{
$
\begin{aligned}
    \tilde{\mathcal{O}}(\frac{\sigma^2\sum_{i=1}^n p_i^2}{\mu\epsilon }+\frac{\sqrt{nLd\sum_{i=1}^n p_i^2 \lambda_i^2}}{\mu \sqrt{\epsilon}}+\frac{L}{\mu})
\end{aligned}
$
}
\end{equation}
iterations of non-uniform D-EF-SGD with the absolute compressor ensures $\mathbb{E}f(\textbf{x}_{final})-f^* \leq \epsilon$, and $\textbf{x}_{final} = \textbf{x}_t$ refers to a version $\textbf{x}_t$ from the sequence $\left\{\textbf{x}_0, \ldots, \textbf{x}_{T-1}\right\}$, selected probabilistically based on $(1-\frac{\mu \gamma}{2})^{-t}$.
\end{theorem}

We predominantly direct our attention to \textbf{Theorem ~\ref{strong convex D-EF-SGD-A}}. Analogous to the analysis of non-uniform D-EF-SGD with the absolute compressor, our focus is constrained to scenarios wherein $\sigma=0$ or during initial training phases. Under these circumstances, the non-uniform D-EF-SGD with the absolute compressor converges at a rate of $\mathcal{O}(\sqrt{\sum_{i=1}^n p_i^2 \lambda_i^2})$.

\subsection{The optimal thresholds in communication-constrained non-IID scenarios}\label{V-B}

We demonstrate the theorem on the optimal $\lambda_i$, \textit{i.e.}, \textbf{Theorem~\ref{theorem-DAGC-A}} and DAGC-A,  We use the Lagrange multiplier method to prove this theorem and more details can be seen in Appendix. 

\begin{theorem}[Conversion from $\lambda$ to $\delta$ and optimal $\lambda_i$]
\label{theorem-DAGC-A}
In the non-IID scenario where the total communication traffic is predetermined and constrained, we have

\begin{equation}
\lambda \propto \frac{1}{\delta}
\end{equation}
and the optimal $\lambda_i$ satisfying

\begin{equation}
\lambda_i=\frac{\Bar{\lambda} P }{n} p_i^{-\frac{2}{3}}, \forall i \in [n],
\end{equation}
where $P=\sum_{i=1}^n p_i^{\frac{2}{3}}$ and $\Bar{\lambda}=\frac{n}{\sum_{i=1}^n \frac{1}{\lambda_i}}$.
\end{theorem}

\begin{remark}
In communication-constrained environments with non-IID datasets, the relationship between $\lambda$ and $\delta$ is different from the previous work \cite{hardthreshold}, which only presents a conversion formula applied to IID scenarios\footnote{The formula, demonstrated in end of the appendix of the work \cite{hardthreshold}, is $\lambda \propto \frac{1}{\sqrt{\delta}}$. They derived this equation by taking $\zeta=0$, equal to IID environments.}.
\end{remark}

\begin{remark}
In IID scenarios, it follows that $\lambda_i = \bar{\lambda}$ for all $i$ due to $p_i = p_j$. This implies that for the absolute compressor,  the optimal compression strategy in IID scenarios is also the uniform compression.
\end{remark}

\subsection{DAGC-A and its implementation in FL}\label{V-C}

We demonstrate the formula for the optimal $\lambda_i$ in \textbf{Theorem~\ref{theorem-DAGC-A}}, based on which we show the pseudo-code of DAGC-A in Algorithm~\ref{algo DAGC-A}. The difference between DAGC-A and DAGC-R is that the time complexity of computation in DAGC-A is only $\mathcal{O}(1)$, and the rest makes no difference in the implementation. The reason for this simplicity is that we make a one-step approximation to the computation of $\lambda_i$ to $\delta_i$, \textit{i.e.,} $\frac{p_i}{p_j} = (\frac{\lambda_i}{\lambda_j})^{-\frac{3}{2}}$. This simplification takes into account the constrained arithmetic in edge computing. We sacrifice some performance to    
reduce the time complexity of DAGC from $\mathcal{O}(n)$ to $\mathcal{O}(1)$.

\begin{algorithm}[t]
    \SetAlgoLined
	\caption{DAGC-A}
	\label{algo DAGC-A}
    \KwIn{ $n$, $p_i$ and the average threshold $\Bar{\lambda}$} 
	\KwOut{thresholds $\lambda_1,\ldots,\lambda_n$}
        $P=\sum_{i=1}^n p_i^{\frac{2}{3}}$\;
        \For{$i = 1, \ldots, n$}{
        \tcc{calculating $\lambda_i$ according to \textbf{Theorem \ref{theorem-DAGC-A}}}
        $\lambda_i=\frac{\Bar{\lambda} P }{n} p_i^{-\frac{2}{3}}$
        }
	\textbf{Return} $\lambda_1,\ldots,\lambda_n$\;
\end{algorithm}
\section{Evaluation Experiments} \label{VI}
The following questions are addressed by this evaluation:

\noindent$\bullet$ If the communication budget is limited and fixed, does DAGC surpass uniform compression in real-world datasets? (Fig.~\ref{fig3} in Sec.~\ref{VI-B})

\noindent$\bullet$ As the size distribution becomes more imbalanced and the compression becomes more aggressive, will DAGC exhibit better performance? (Table~\ref{table:3} in Sec.~\ref{VI-C} and Table~\ref{table:4} in Sec.~\ref{VI-E})

We showcase the superior performance of DAGC in both real-world non-IID and artificially partitioned non-IID datasets, particularly when confronted with highly imbalanced size distribution and constrained communication.

\begin{table*}[!th]
\caption{Summary of the experiment settings used in this work.}
\label{table:1}
\centering
\begin{threeparttable}
\begin{tabular}{cccccccc}
   \toprule
   Task & Model  & Dataset & non-IID type & Quality metric &  Training iterations & Experiment Section \\
   \midrule
   \multirow{5}{*}{Image Classification} &   Logistic
   & FMNIST \cite{fmnist}  & Artificially  & \multirow{5}{*}{Top-1 Accuracy} & $5,000$ & Sec.~\ref{III}, \ref{VI} \\
    &   CNN
   & CIFAR-10
   \cite{cifar10} & Artificially  & &  $10,000$ & Sec. \ref{VI}\\
   &    VGG11s & Flickr \cite{niid2020icml} & Real-world && $10,000$ & Sec. \ref{VI} \\
   &    VGG11 &  CIFAR-100 \cite{krizhevsky2009learning} & Artificially && $50,000$ & Sec. \ref{VI} \\
   &   ResNet18
   & CIFAR-10 
    & Artificially  & &  $10,000$ & Sec. \ref{VI}\\
   \midrule
   Speech Recognition &   LSTM &  SCs \cite{warden2018speech} & Artificially & Top-1 Accuracy  &  $10,000$ & Sec. \ref{III}, \ref{VI} \\
   \bottomrule
   
\end{tabular}
\end{threeparttable}
\label{benchmark}
\end{table*}
 
\subsection{Experimental settings}\label{VI-A}

\noindent \textbf{Environment:} 
The experiments are conducted on an Ubuntu $18.04.6$ LTS server environment. The server is equipped with an Intel Xeon Silver $4210$ CPU @$2.20$GHz and $4$ Nvidia GeForce GTX $3090$ GPUs, each with $24$GB memory. Python version $3.8.12$ is utilized, along with various libraries compatible with this Python version. For machine learning purposes, PyTorch $1.11.0$ with CUDA $11.3$ is employed as the primary toolkit.

\noindent \textbf{Non-IID type:}
We run the experiments in two non-IID types:

\noindent$\bullet$ \textit{Artificial non-IID data partition:} 
In order to simulate label imbalance, we assign a portion of samples from each label to individual workers based on the Dirichlet distribution. The concentration parameter is set to $0.5$. This partitioning strategy is widely adopted for generating non-IID data \cite{li2020practical,niidbench,wang2020federated}. 

\begin{figure*}[t]
	\centering
	\subfloat[The label distribution for Flickr]{\label{fig3:a}
		\begin{minipage}[b]{0.38\linewidth}
			\vfill
			\includegraphics[width=1\linewidth]{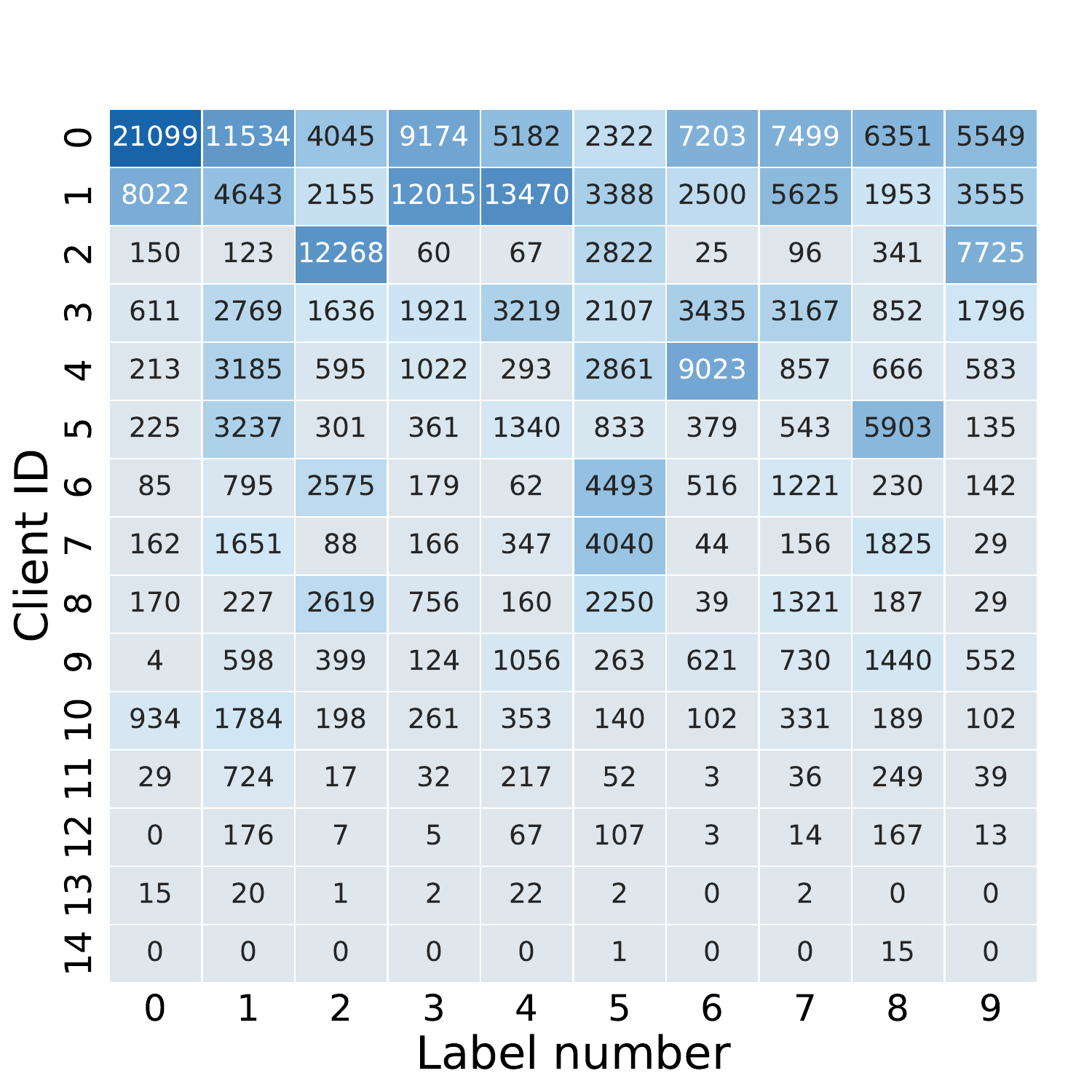}
		\end{minipage}
	}
	\begin{minipage}[b]{0.62\linewidth}
		\subfloat[$\Bar{\delta}=10\%$]{\label{fig3:b}
			\includegraphics[width=0.3\linewidth]{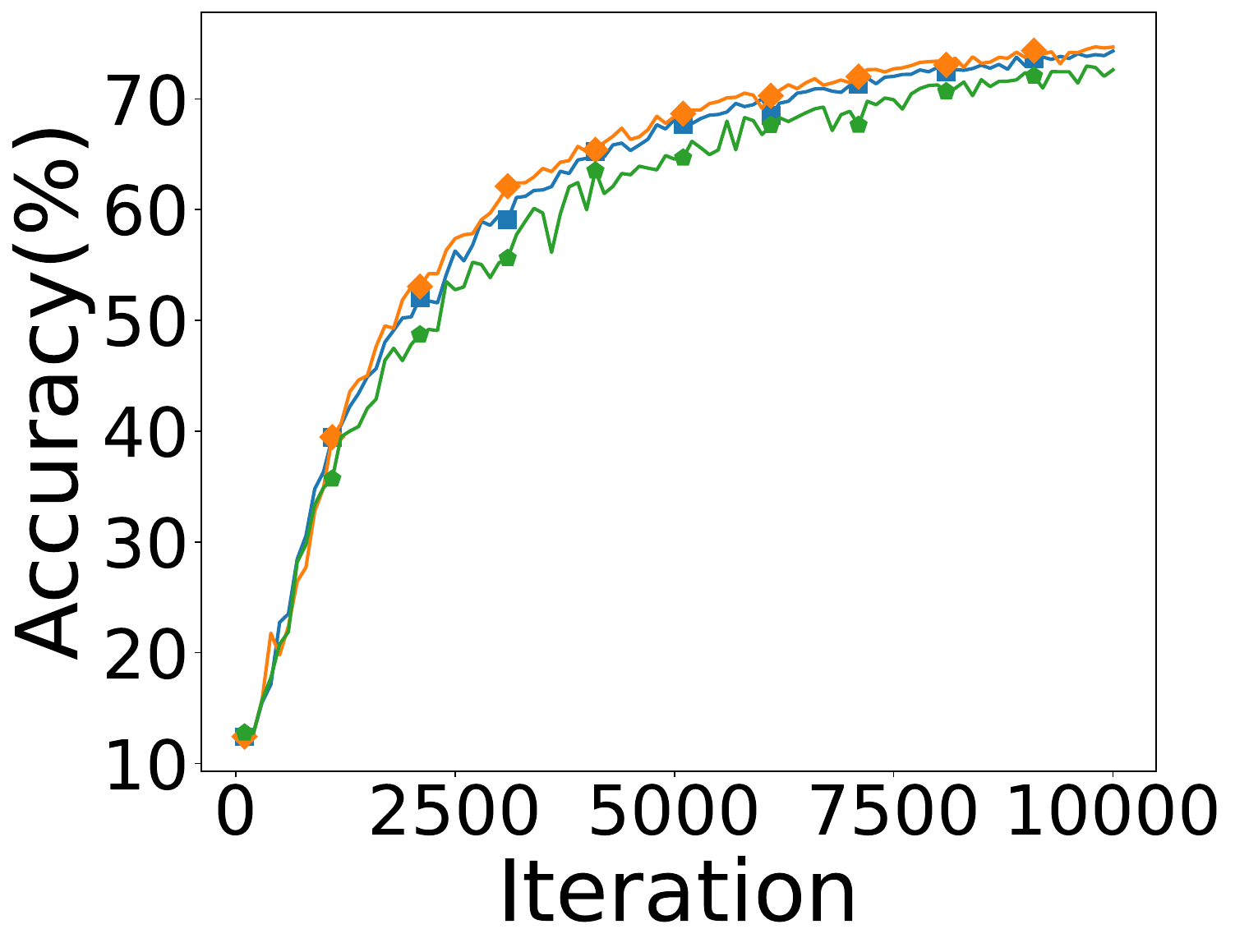}
		}
		\subfloat[$\Bar{\delta}=1\%$]{\label{fig3:c}
			\includegraphics[width=0.3\linewidth]{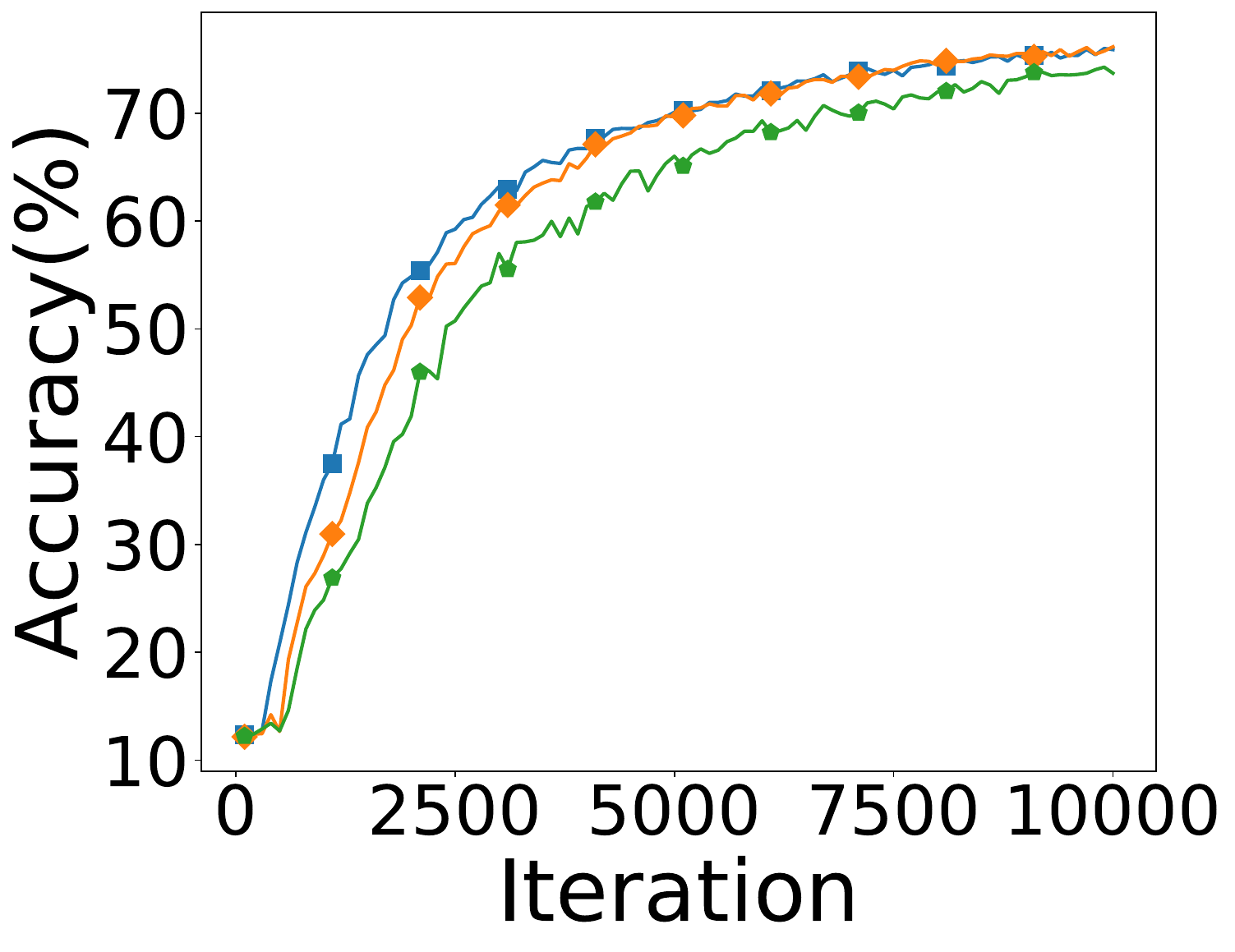}
		}
		\subfloat[$\Bar{\delta}=0.1\%$]{\label{fig3:d}
			\includegraphics[width=0.3\linewidth]{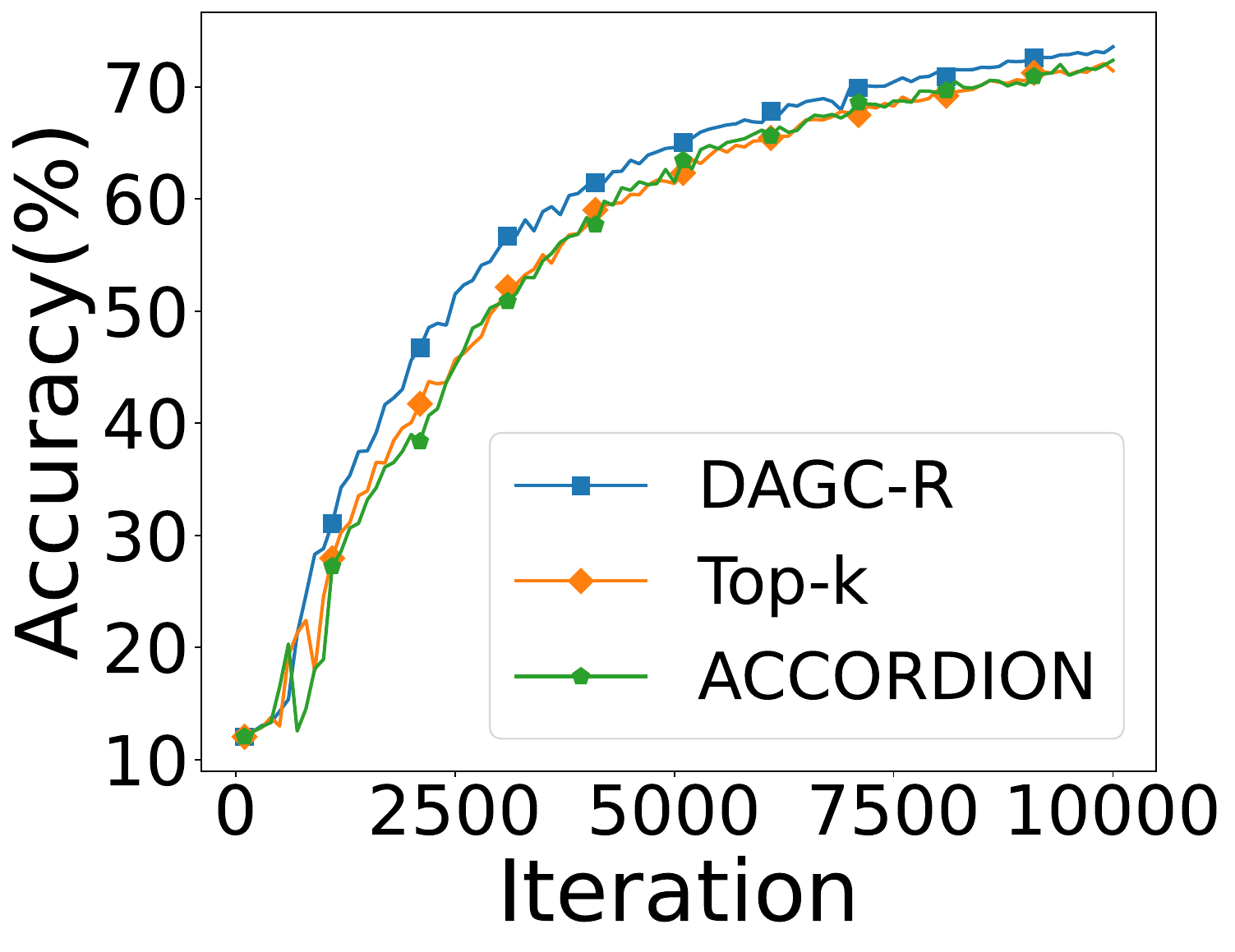}
		}
        
		\subfloat[$\Bar{\lambda}=5.0 \times 10^{-4} $]{\label{fig3:e}
			\includegraphics[width=0.3\linewidth]{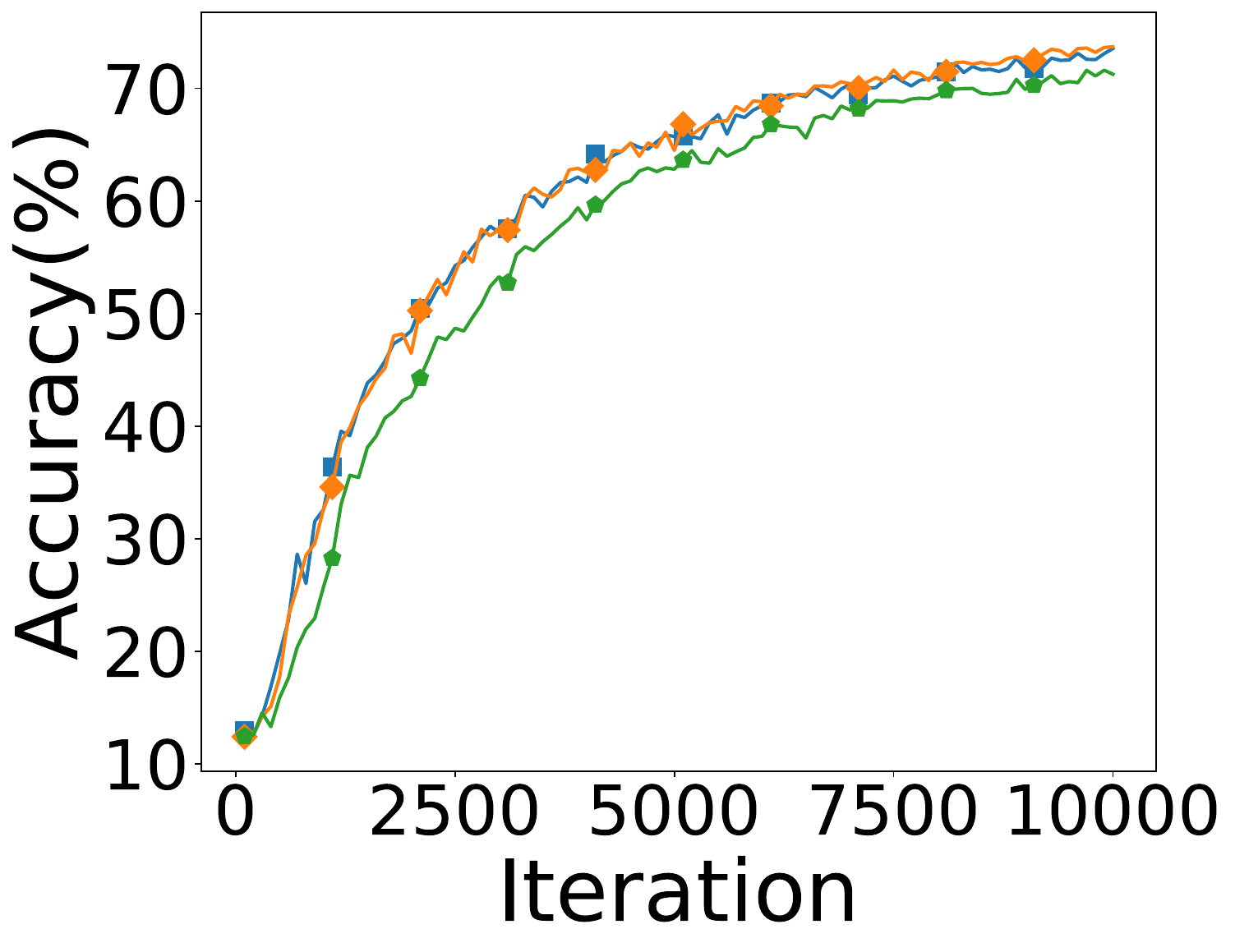}
		}
		\subfloat[$\Bar{\lambda}=5.0 \times 10^{-3} $]{\label{fig3:f}
			\includegraphics[width=0.3\linewidth]{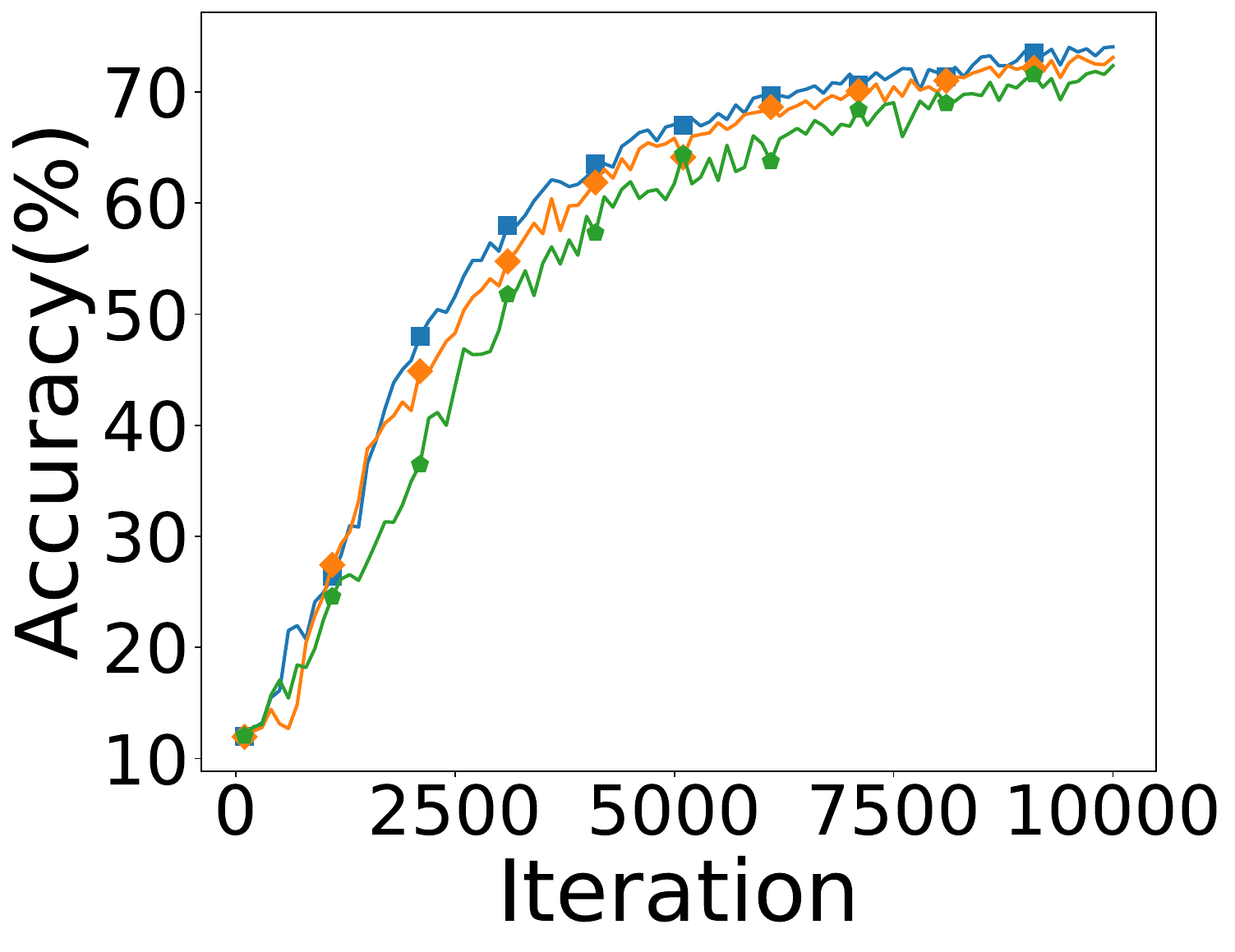}
		}
		\subfloat[$\Bar{\lambda}=5.0 \times 10^{-2} $]{\label{fig3:g}
			\includegraphics[width=0.3\linewidth]{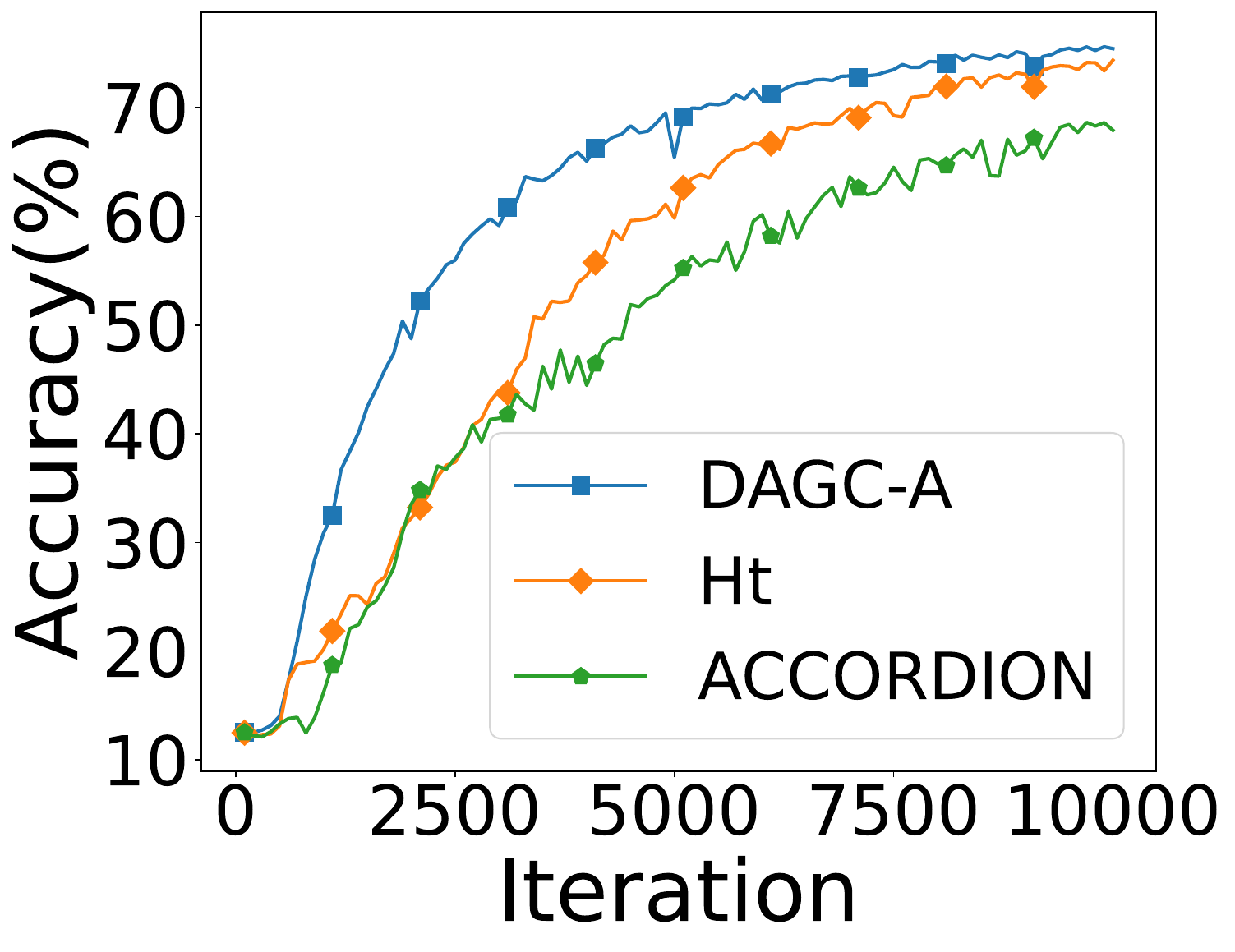}
	}
    \end{minipage}
\caption{The label distribution for Flickr (a) and training curves (Accuracy vs. Iterations) for VGG11s@Flickr under the relative compression ((b)-(d)) and the absolute compression ((e)-(g)) on different compression levels (left to right). DAGC outperforms other uniform compression strategies facing limited communication under the fixed budget.}
\label{fig3}
\end{figure*}

\noindent$\bullet$  \textit{Real-world datasets:} We utilize the Flickr \cite{niid2020icml} dataset as our real-world dataset. The dataset is downloaded from \texttt{https://doi.org/10.5281/zenodo.3676081} and the images are divided based on the subcontinent they belong to. After excluding damaged images and accounting for network limitations that prevented the download of certain images, we have a total of 15 workers. The data distribution is illustrated in Fig.~\ref{fig3:a}.

\noindent \textbf{Experiment tasks:}
The experimental settings in this study encompass four different types, involving tasks related to image classification and speech recognition in Tab.~\ref{table:1}. The Convolutional Neural Network (CNN) employed consists of four layers, as mentioned earlier \cite{fed_learning}. VGG11 \cite{vgg} consists of 11 layers (including 8 convolutional layers and 3 fully-connected layers), using a continuous 3x3 small convolutional kernel. VGG11s  is a simplified version of the VGG11 \cite{sbc}. ResNet18 \cite{resnet18} is a deep convolutional neural network consisting of 18 convolutional layers and residual blocks, each of which mitigates the gradient vanishing problem by constant mapping. The LSTM model utilized has $2$ hidden layers with a size of $128$. In order to mitigate the accuracy loss caused by non-IID, all of these models exclude the batch normalization layer \cite{niid2020icml}.

We denote A@B as the task, which uses the B datasets to train A model. For tasks CNN@CIFAR-10, VGG11s@Flickr, VGG11@CIFAR-100, and ResNet18@CIFAR-10, the learning rates are $0.01$, $0.1$, $0.01$, and $0.01$, respectively, and the batch sizes are $32$ for all tasks. For the Speech Recognition task, we utilize the Speech Commands dataset \cite{warden2018speech} (referred to as SCs). From SCs, we select the $10$ categories with the highest number of samples. Specifically, we extract $4,000$ samples from each of these categories, with $3,000$ samples allocated for training purposes and the remaining $1,000$ samples reserved for testing. The batch size is $8$ and the step size is $0.1$ in LSTM@SCs.

\noindent \textbf{Baselines:} In our comparative analysis, DAGC is evaluated against established uniform compression strategies. For DAGC-R, Top-$k$ and ACCORDION \cite{accordion} serve as the baselines. In the case of DAGC-A, the hard-threshold (denoted as Ht) and ACCORDION are the selected baselines. The compression ratio of Top-$k$ is set to $\Bar{\delta}$ and the threshold of Ht is  $\Bar{\lambda}$. ACCORDION is the state-of-the-art sparsified adaptive gradient compression algorithm, which compresses the gradient using aggressive compression in the critical regime and conservative compression if not. Specifically, within the relative compressor, the aggressive (conservative) compression ratio of ACCORDION is set to $\delta_{\min}$ ($\delta_{\max}$). Conversely, under the absolute compression, the aggressive compression threshold is set to $\lambda_{\max}$.

\noindent \textbf{The number of workers and the worker size:}
This experimental setting is specifically for generating artificial non-IID data partitions. We set the number of workers equal to $10$. The worker size does not undergo a dichotomous division (as used in Sec.~\ref{III} for simplicity). Instead, $p_i$ is an arithmetic series. To increase the randomness of the series and to better match real-world datasets, we incorporate a Dirichlet distribution (with a concentration parameter of $0.5$) into $p_i$, $i \in [2,n-1]$. This results in an approximate arithmetic series that remains descending order.
This approach allows us to generate artificially non-IID datasets with different worker sizes in Federated Learning \cite{niidbench,wang2020tackling}. To measure the imbalance of datasets, we define the skew ratio (abbreviated as SR) as $p_1/p_n$.

\subsection{DAGC-R in real-world non-IID scenarios}\label{VI-B}
The experimental results indicate that the performance of DAGC-R surpasses Top-$k$ with uniform compression and ACCORDION in real-world scenarios with fixed communication volumes.

Fig.~\ref{fig3:a} displays the data distribution of Flickr \cite{niid2020icml} (sliced by subcontinent), which has a skew ratio of $4,997$ ($\approx\frac{79958}{16} $). We find out the $10$ categories with the highest number of labels in all the images and select these $10$ categories of images from $15$ workers as the training dataset.

From Fig.~\ref{fig3:b}, it can be seen that at an average compression rate $\Bar{\delta} = 10\%$, DAGC-R converges almost as fast as Top-$k$ and both are slightly faster than ACCORDION.

Fig.~\ref{fig3:c} shows that DAGC-R converges faster than Top-$k$ in the early stages, and then gradually equalizes with Top-$k$ later on. DAGC-R has always been faster than ACCORDION. 

The training accuracy curve in Fig.~\ref{fig3:d} shows that at $\Bar{\delta} = 0.1\%$, DAGC-R consistently has a superior performance compared to Top-$k$ and ACCORDION. DAGC-R achieves the same accuracy ($70\%$) with $16.65\%$ and $13.46\%$ fewer iterations (from $8,410$ iterations and $8,100$ iterations to $7,010$ iterations) relative to Top-$k$ and ACCORDION, respectively.

Overall, DAGC-R outperforms Top-$k$ with uniform compression and ACCORDION in the normal compression interval with fixed communication volumes.

\begin{table}[t]
\centering
\caption{Accuracy of different relative gradient compression algorithms under different SR and average relative compression ratios $\Bar{\delta}$. Increasing SR indicates a greater imbalance among the size distribution of datasets. $\Bar{\delta}$ quantifies the extent of communication limitations. Numbers demonstrate that DAGC-R surpasses the performance of the uniform compression on all tasks. The superiority is particularly notable in environments where worker size distribution is highly uneven and communication bandwidth is constrained.}
\label{table:3}
\scalebox{0.9}{
\begin{tabular}{|c|c|c|c|c|c|}
\hline
\makecell[c]{Model\\@Dataset}& SR & 
$\Bar{\delta}$ & \textbf{DAGC-R} & Top-$k$ & ACCORDION\\
\hline
\hline
\multirow{9}*{\makecell[c]{CNN\\@CIFAR-10}} & \multirow{3}*{$10$} & $10\%$ & $70.02\% $ & $\mathbf{70.14}\% $ & $69.77\% $ \\
\cline{3-6}
& & $1\% $& $69.66\% $ & $\mathbf{69.68}\% $ & $69.60\% $ \\
\cline{3-6}
& & $0.1\% $ & $\mathbf{68.85}\% $ & $68.72\% $ & $68.77\% $ \\
\cline{2-6}
& \multirow{3}*{$100$} & $10\% $& $68.68\% $ & $\mathbf{68.79}\% $ & $68.59\% $ \\
\cline{3-6}
& & $1\%$ & $\mathbf{69.25}\% $ & $68.87\% $ & $68.46\% $\\
\cline{3-6}
& & $0.1\%$ & $\mathbf{69.29}\% $ & $67.36\% $ & $67.93\% $\\
\cline{2-6}
&\multirow{3}*{$1,000$} & $10\% $& $68.00\% $ & $\mathbf{68.12}\% $ & $67.89\% $ \\
\cline{3-6}
& & $1\%$ & $\mathbf{68.06}\% $ & $67.28\% $ & $66.49\% $ \\
\cline{3-6}
& & $0.1\%$ & $\mathbf{68.25}\% $ & $67.35\% $ & $67.75\% $ \\
\hline

\multirow{9}*{\makecell[c]{LSTM\\@SCs}} & \multirow{3}*{$10$} & $10\%$ & $78.40\% $ & $\mathbf{78.53}\% $ & $77.93\% $ \\
\cline{3-6}
& & $1\% $& $\mathbf{77.10}\% $ & $76.37\% $ & $76.87\% $ \\
\cline{3-6}
& & $0.1\% $ & $\mathbf{77.07}\% $ & $76.23\% $ & $76.93\% $ \\
\cline{2-6}
& \multirow{3}*{$100$} & $10\% $& $75.17\% $ & $\mathbf{75.50}\% $ & $74.70\% $ \\
\cline{3-6}
& & $1\%$ & $\mathbf{74.60}\% $ & $73.53\% $ & $73.87\% $\\
\cline{3-6}
& & $0.1\%$ & $\mathbf{73.33}\% $ & $71.53\% $ & $70.70\% $\\
\cline{2-6}
&\multirow{3}*{$1,000$} & $10\% $& $\mathbf{73.03}\% $ & $72.87\% $ & $71.43\% $ \\
\cline{3-6}
& & $1\%$ & $\mathbf{72.67}\% $ & $71.17\% $ & $72.03\% $ \\
\cline{3-6}
& & $0.1\%$ & $\mathbf{71.80}\% $ & $70.27\% $ & $71.13\% $ \\

\hline

\multirow{9}*{\makecell[c]{Logistic\\@FMNIST}} & \multirow{3}*{$10$} & $10\%$ & $\mathbf{83.44}\%$ & $83.38\%$ & $83.41\%$ \\
\cline{3-6}
& & $1\%$ & $\mathbf{83.36}\%$ & $83.21\%$ & $83.32\%$ \\
\cline{3-6}
& & $0.1\%$ & $\mathbf{83.18}\%$ & $83.09\%$ & $83.15\%$ \\
\cline{2-6}
& \multirow{3}*{$100$} &$10\%$ & $83.35\%$ & $83.38\%$ & $\mathbf{83.41}\%$ \\
\cline{3-6}
& & $1\%$ & $\mathbf{83.26}\%$ & $83.23\%$ & $83.18\%$ \\
\cline{3-6}
& & $0.1\%$ & $\mathbf{83.13}\%$ & $82.92\%$ & $83.10\%$ \\
\cline{2-6}
& \multirow{3}*{$1,000$} & 10\% & $\mathbf{83.17}\%$ & $\mathbf{83.17}\%$ & $82.86\%$ \\
\cline{3-6}
& & $1\%$ & $\mathbf{83.24}\%$ & $83.05\%$ & $83.16\%$ \\
\cline{3-6}
& & $0.1\%$ & $\mathbf{83.06}\%$ & $82.95\%$ & $82.90\%$ \\

\hline
\end{tabular}
}
\end{table}

\begin{table}[!t]
\centering
\caption{Accuracy of different relative compressors under  different numbers of workers $n$ in CNN@CIFAR-10 under SR=$100$ and $\Bar{\delta}=0.1\%$.}
\label{table:worker size:DAGC-R}
\scalebox{1.2}{
\begin{tabular}{|c|c|c|c|}
\hline
$n$ & \textbf{DAGC-R} & Top-$k$ & ACCORDION \\
\hline
\hline
$5$  & $\mathbf{67.68}\%$ & $67.02\%$ & $67.03\%$ \\
\cline{1-4}
$10$ & $\mathbf{69.29}\%$ & $68.87\%$ & $68.46\%$ \\
\cline{1-4}
$20$ & $\mathbf{68.94}\%$ & $68.55\%$ & $68.70\%$ \\
\cline{1-4}
$50$ & $\mathbf{69.33}\%$ & $68.53\%$ & $68.60\%$ \\
\hline
$100$ & $\mathbf{69.77}\%$ & $69.26\%$ & $69.19\%$ \\
\hline
$200$ & $\mathbf{69.82}\%$ & $69.24\%$ & $69.79\%$ \\
\hline
\end{tabular}
}
\end{table}

\subsection{DAGC-R in artificially partitioned non-IID scenarios}\label{VI-C}

The detailed experimental results presented in  illustrate the outperformance of DAGC-R compared to Top-$k$ and ACCORDION in the case of highly skewed datasets.

\noindent \textbf{Comparison of different skew ratios:} The superiority of DAGC-R in both CNN@CIFAR-10 and LSTM@SCs tasks becomes more obvious as the skew ratio increases. DAGC-R, Top-$k$ and ACCORDION perform similarly when $\text{SR}=10$. However, when SR increases to $1,000$, the accuracy of DAGC-R is higher than Top-$k$ and ACCORDION.

\noindent \textbf{Comparison of different $\Bar{\delta}$:} DAGC-R is always the best in both tasks regardless of the skew ratio, when $\Bar{\delta}=0.1\%$. This suggests that DAGC-R is suitable for situations where communications are extremely limited.

\noindent \textbf{Comparison of different numbers of workers:}
In Table~\ref{table:worker size:DAGC-R}, we compare the performance of several compression algorithms under different $n$. Even under $n = 200$, DAGC-R still shows better performance than other algorithms, which indicates the scalability of DAGC-R.

\subsection{DAGC-A in real-world non-IID scenarios}\label{VI-D}

The experimental results show that DAGC-A outperforms hard-threshold and ACCORDION in the real-world non-IID dataset. The advantage over hard-threshold and ACCORDION becomes more pronounced as compression becomes more aggressive.

Fig.~\ref{fig3:e} and Fig.~\ref{fig3:f} illustrate that in the conservative case of compression, DAGC-A and the hard-threshold perform similarly but both outperform ACCORDION. 

Fig.~\ref{fig3:g} shows that under very aggressive compression, DAGC-A converges much faster than hard-threshold and ACCORDION, especially in the first half of the training stage exhibiting a very clear advantage. DAGC-A ultimately saves $25.43\%$ of iterations (from $6,960$ iterations to $5,190$ iterations) over hard-threshold to reach $70\%$ accuracy, while ACCORDION does not reach the same accuracy until the end of training.

\subsection{DAGC-A in artificially partitioned non-IID scenarios}\label{VI-E}

The experimental results show that DAGC-A performs better than Ht and ACCORDION in extremely communication-constrained situations and datasets with high skew ratios. The detailed experimental results are in Table~\ref{table:4}.

\noindent \textbf{Comparison of different skew ratios:} DAGC-A behaves more prominently when the skew ratio is relatively large. In the CNN@CIFAR-10 task, fixing $\Bar{\lambda} = 0.05 $, the training accuracy of DAGC-A is $1.35\%$, $6.54\%$, and $7.72\%$ higher than that of Ht for the skew ratio of $10, 100, 1,000$, respectively. In the LSTM@SCs task, DAGC-A always performs best when $\text{SR}=1,000$, while it only performs best when $\text{SR}=10$ with average $\Bar{\lambda} = 0.0005 $.

\noindent \textbf{Comparison of different $\Bar{\lambda}$:} When the average lambda is larger, which means that in more aggressive compression, DAGC-A always achieves the highest accuracy after the same number of iterations. Keeping the skew ratio constant, the advantage of DAGC-A in both CNN@CIFAR-10 and LSTM@SCs tasks becomes more obvious as $\Bar{\lambda}$ increases.

\noindent \textbf{Comparison of different numbers of workers:} As $n$ increases, the accuracy improvement decreases. This is because the larger worker size increases the number of input samples per iteration, thus shortening the training iterations. However, DAGC-A is still superior to other compression strategies under varying $n$.

\begin{table}[!t]
\centering
\caption{Accuracy of different absolute gradient compression algorithms under different SR and average absolute compression thresholds $\Bar{\lambda}$. A larger $\Bar{\lambda}$ represnets a more aggressive compression. DAGC-A also surpasses the performance of the uniform absolute compressors in communication-constrained non-IID scenarios.}
\label{table:4}
\scalebox{0.9}{
\begin{tabular}{|c|c|c|c|c|c|}
\hline
\makecell[c]{Model\\@Dataset}& SR & 
$\Bar{\lambda}$ & \textbf{DAGC-A} & Ht & ACCORDION\\
\hline
\hline
\multirow{9}*{\makecell[c]{CNN\\@CIFAR-10}} & \multirow{3}*{$10$} & $0.0005$ & $69.79\% $ & $69.73\% $ & $\mathbf{69.89}\% $ \\
\cline{3-6}
& & $0.005$& $\mathbf{68.52}\% $ & $68.06\% $ & $68.39\% $ \\
\cline{3-6}
& & $0.05$ & $\mathbf{63.69}\% $ & $62.34\% $ & $61.80\% $ \\
\cline{2-6}
& \multirow{3}*{$100$} & $0.0005$ & $69.12\% $ & $\mathbf{69.31}\% $ & $68.37\% $ \\
\cline{3-6}
& & $0.005$ & $\mathbf{68.25}\% $ & $67.59\% $ & $67.65\% $\\
\cline{3-6}
& & $0.05$ & $\mathbf{63.47}\% $ & $56.93\% $ & $60.33\% $\\
\cline{2-6}
&\multirow{3}*{$1,000$} & $0.0005$ & $68.04\% $ & $\mathbf{68.12}\% $ & $67.28\% $ \\
\cline{3-6}
& & $0.005$ & $ \mathbf{67.27}\% $ & $66.58\% $ & $66.87\% $ \\
\cline{3-6}
& & $0.05$ & $\mathbf{64.17}\% $ & $56.45\% $ & $63.54\% $ \\
\hline

\multirow{9}*{\makecell[c]{LSTM\\@SCs}} & \multirow{3}*{$10$} & $0.0005$ & $77.63\% $ & $\mathbf{78.03}\% $ & $76.60\% $ \\
\cline{3-6}
& & $0.005$ & $76.60\% $ & $\mathbf{76.90}\% $ & $76.73\% $ \\
\cline{3-6}
& & $0.05$ & $\mathbf{75.27}\% $ & $73.97\% $ & $73.43\% $ \\
\cline{2-6}
& \multirow{3}*{$100$} & $0.0005$ & $75.10\% $ & $74.83\% $ & $\mathbf{75.40}\% $ \\
\cline{3-6}
& & $0.005$ & $\mathbf{74.13}\% $ & $73.93\% $ & $72.60\% $\\
\cline{3-6}
& & $0.05$ & $\mathbf{73.37}\% $ & $70.90\% $ & $72.13\% $\\
\cline{2-6}
&\multirow{3}*{$1,000$} & $0.0005$ & $\mathbf{72.13}\% $ & $72.10\% $ & $71.10\% $ \\
\cline{3-6}
& & $0.005$ & $\mathbf{71.97}\% $ & $70.57\% $ & $70.70\% $ \\
\cline{3-6}
& & $0.05$ & $\mathbf{70.17}\% $ & $67.53\% $ & $68.70\% $ \\
\hline

\multirow{9}*{\makecell[c]{Logistic\\@FMNIST}} & \multirow{3}*{$10$} & $0.0005$ & $83.31\%$ & $\mathbf{83.35}\%$ & $83.15\%$ \\
\cline{3-6}
& & $0.005$ & $83.26\%$ & $83.19\%$ & $\mathbf{83.30}\%$ \\
\cline{3-6}
& & $0.05$ & $\mathbf{82.99}\%$ & $82.65\%$ & $82.97\%$ \\
\cline{2-6}
& \multirow{3}*{$100$} & $0.0005$ & $\mathbf{83.29}\%$ & $83.16\%$ & $83.20\%$ \\
\cline{3-6}
& & $0.005$ & $\mathbf{83.24}\%$ & $83.20\%$ & $83.17\%$ \\
\cline{3-6}
& & $0.05$ & $\mathbf{83.04}\%$ & $82.50\%$ & $83.04\%$ \\
\cline{2-6}
& \multirow{3}*{$1,000$} & $0.0005$ & $\mathbf{83.19}\%$ & $\mathbf{83.19}\%$ & $82.93\%$ \\
\cline{3-6}
& & $0.005$ & $\mathbf{83.18}\%$ & $83.12\%$ & $82.82\%$ \\
\cline{3-6}
& & $0.05$ & $\mathbf{83.14}\%$ & $82.10\%$ & $83.05\%$ \\

\hline
\end{tabular}
}
\end{table}

\begin{table}[!t]
\centering
\caption{Accuracy of different absolute compression algorithms with varying numbers of workers $n$ in CNN@CIFAR-10 under SR=$100$ and $\Bar{\lambda}=0.05$.}
\label{table:worker size:DAGC-A}
\scalebox{1.2}{
\begin{tabular}{|c|c|c|c|}
\hline
$n$ & \textbf{DAGC-A} & Ht & ACCORDION \\
\hline
\hline
$5$  & $\mathbf{61.68}\%$ & $56.93\%$ & $59.12\%$ \\
\cline{1-4}
$10$ & $\mathbf{63.47}\%$ & $56.93\%$ & $60.33\%$ \\
\cline{1-4}
$20$ & $\mathbf{64.21}\%$ & $60.30\%$ & $64.09\%$ \\
\cline{1-4}
$50$ & $\mathbf{65.96}\%$ & $63.52\%$ & $65.10\%$ \\
\cline{1-4}
$100$ & $\mathbf{66.76}\%$ & $63.34\%$ & $66.39\%$ \\
\cline{1-4}
$200$ & $\mathbf{68.15}\%$ & $65.14\%$ & $66.41\%$ \\
\hline
\end{tabular}
}
\end{table}

\begin{figure}[t]
\centering
\subfloat[DAGC-R]{
\includegraphics[width=0.45\linewidth]{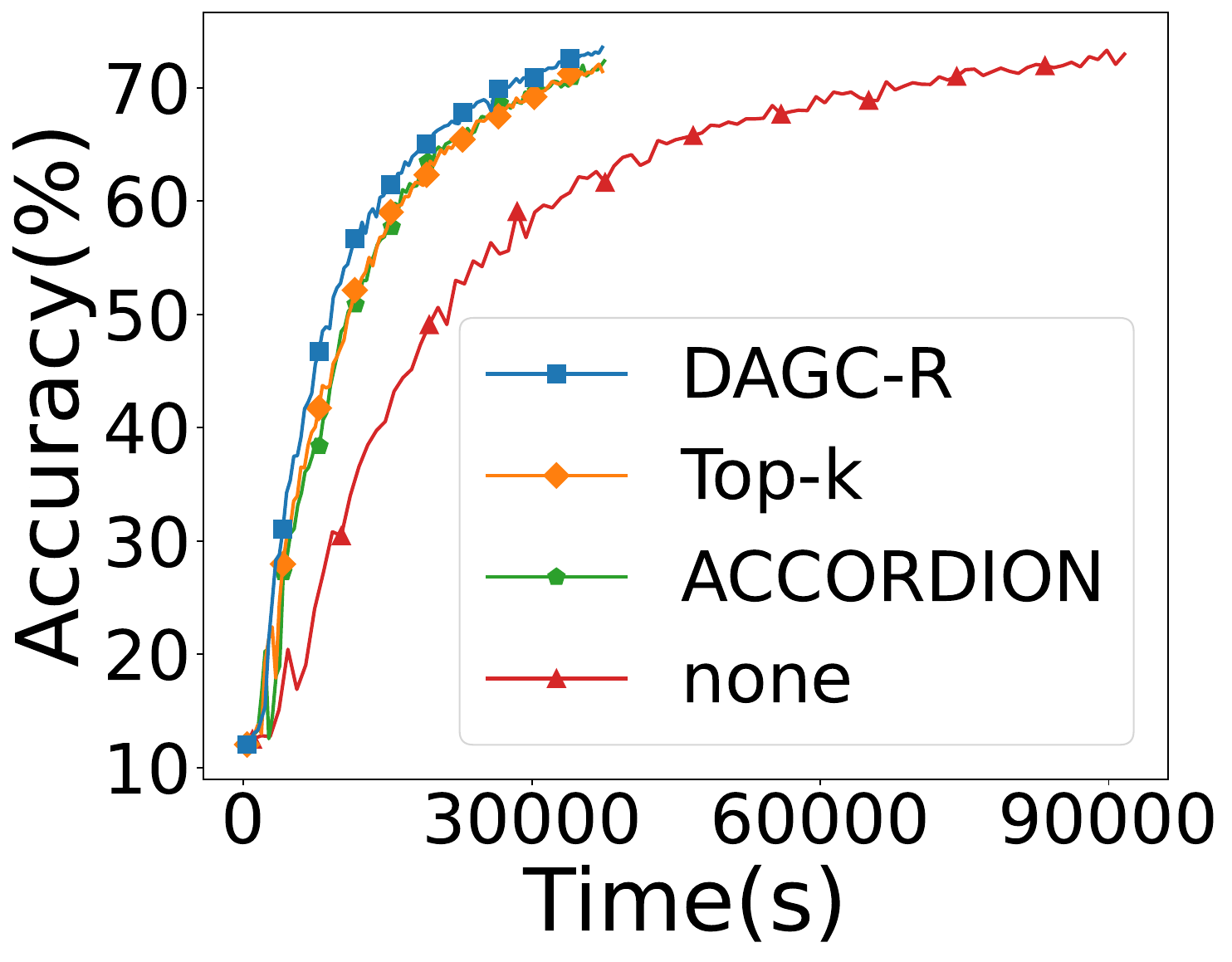}\label{fig:4:a}}
\hspace{0.01\linewidth}
\subfloat[DAGC-A]{\includegraphics[width=0.45\linewidth]{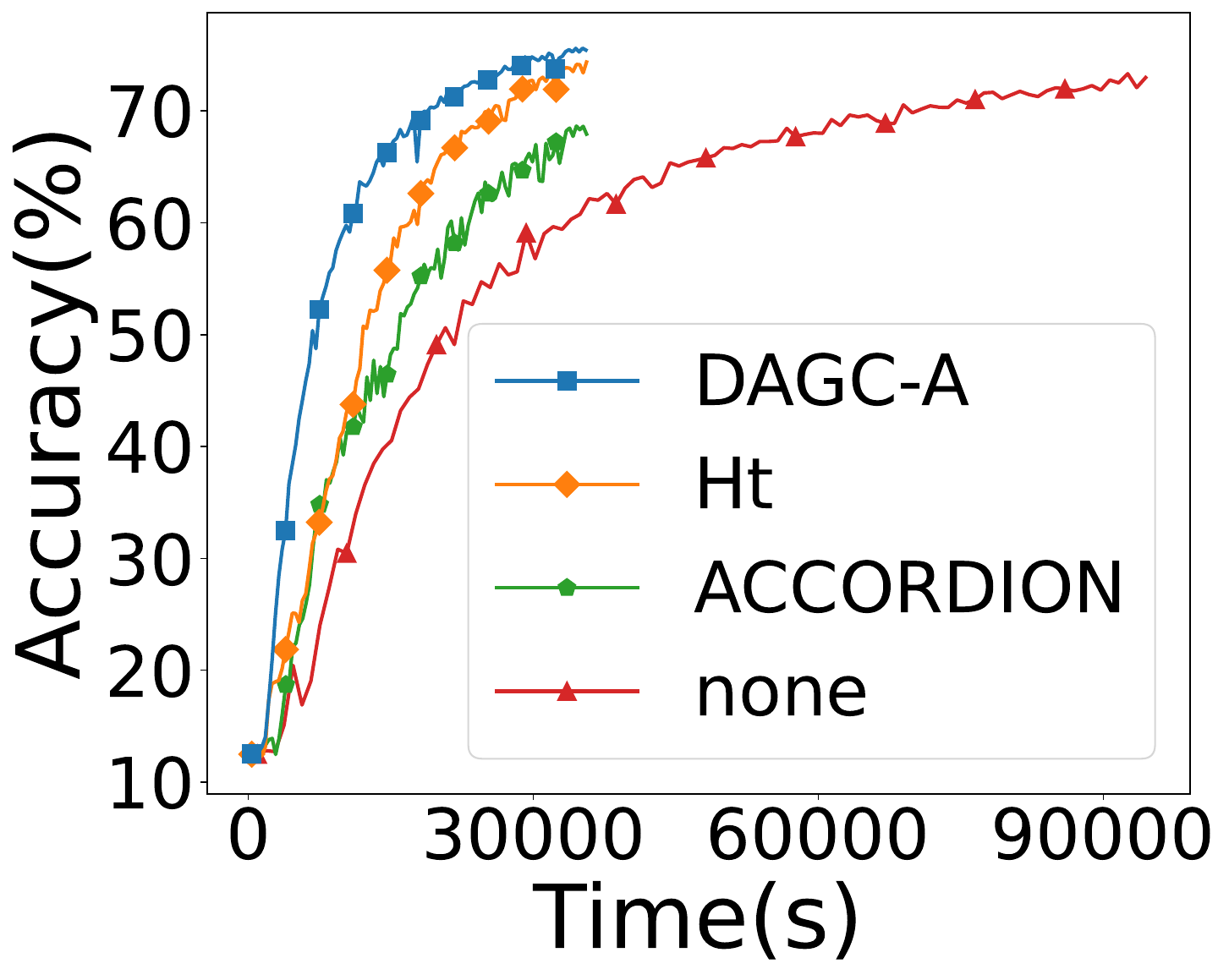}\label{fig:4:b}}
\caption{The training curves (Accuracy vs. Time) for VGG11s@Flickr under the relative compression (a) and the absolute compression (b). DAGC outperforms other compression strategies, and the training is without compression.}
\label{fig:4}
\end{figure}

\begin{figure}[t]
\captionsetup[subfigure]{justification=centering}
\centering
\subfloat[DAGC-R \\ ResNet18@CIFAR-10]{
\includegraphics[width=0.45\linewidth]{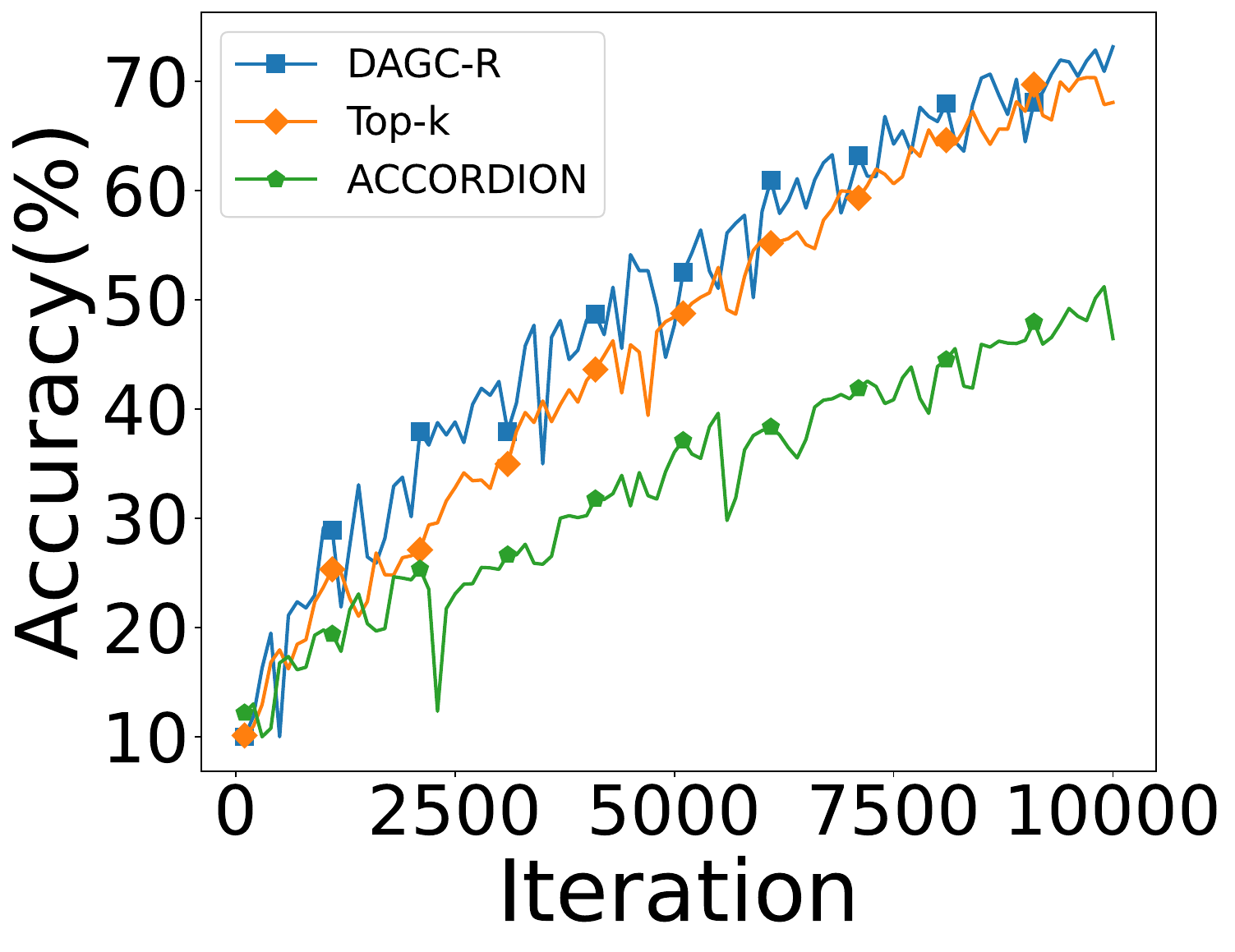}\label{fig:5:a}}
\hspace{0.01\linewidth}
\subfloat[DAGC-A \\ ResNet18@CIFAR-10]{
\includegraphics[width=0.45\linewidth]{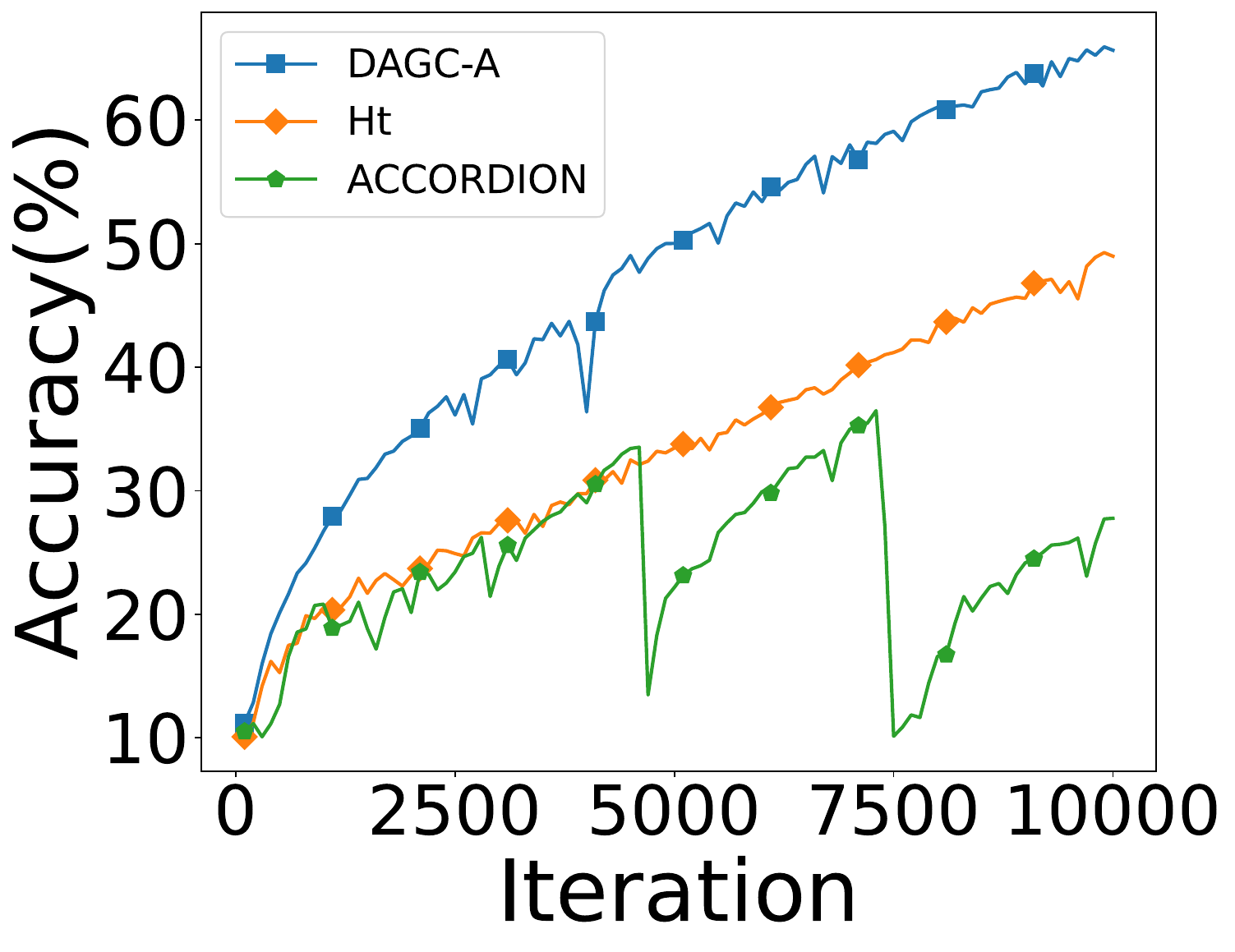}\label{fig:5:b}}

\vspace{0.01\linewidth}

\subfloat[DAGC-R \\ VGG11@CIFAR-100]{
\includegraphics[width=0.45\linewidth]{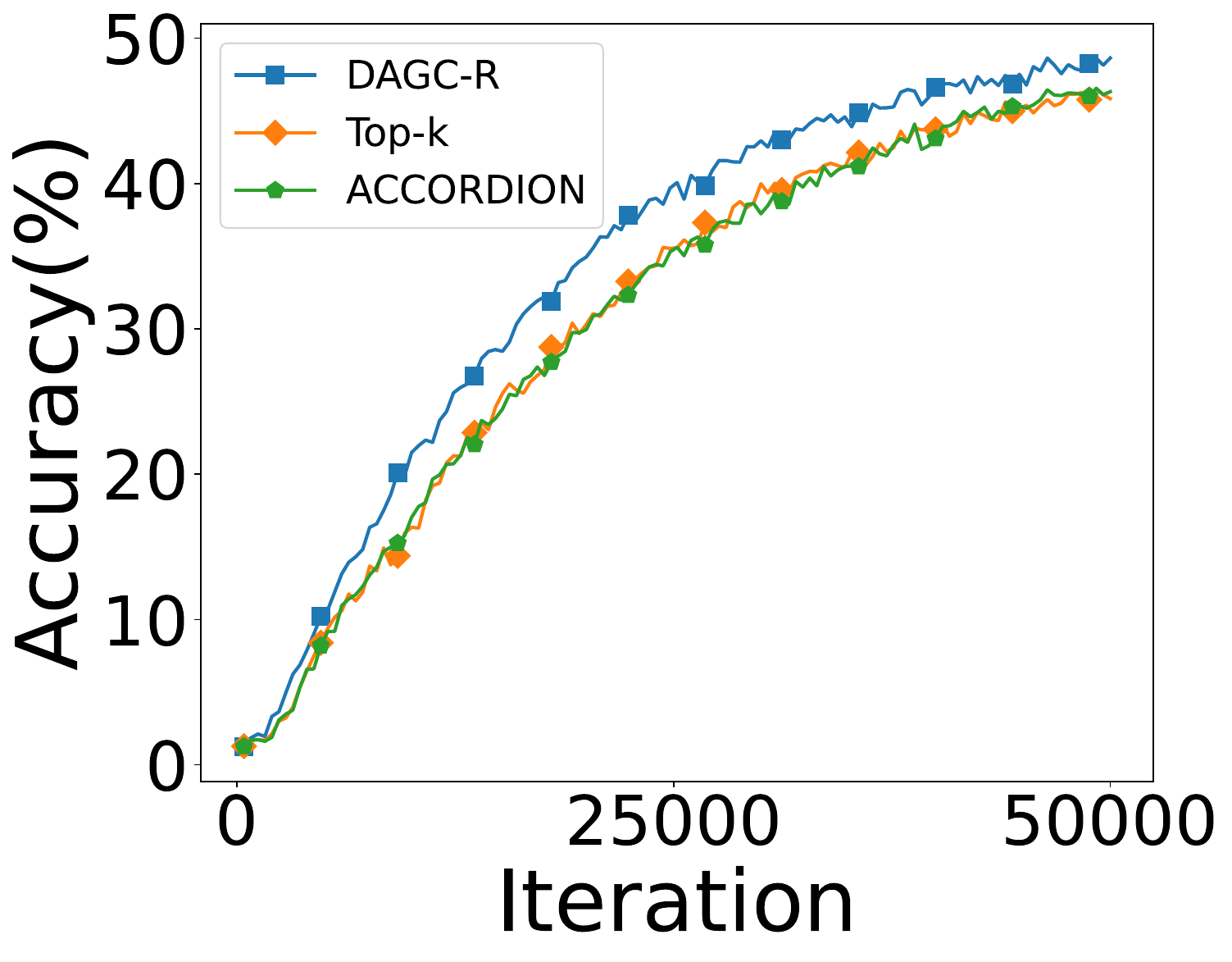}\label{fig:5:c}}
\hspace{0.01\linewidth}
\subfloat[DAGC-A \\ VGG11@CIFAR-100]{
\includegraphics[width=0.45\linewidth]{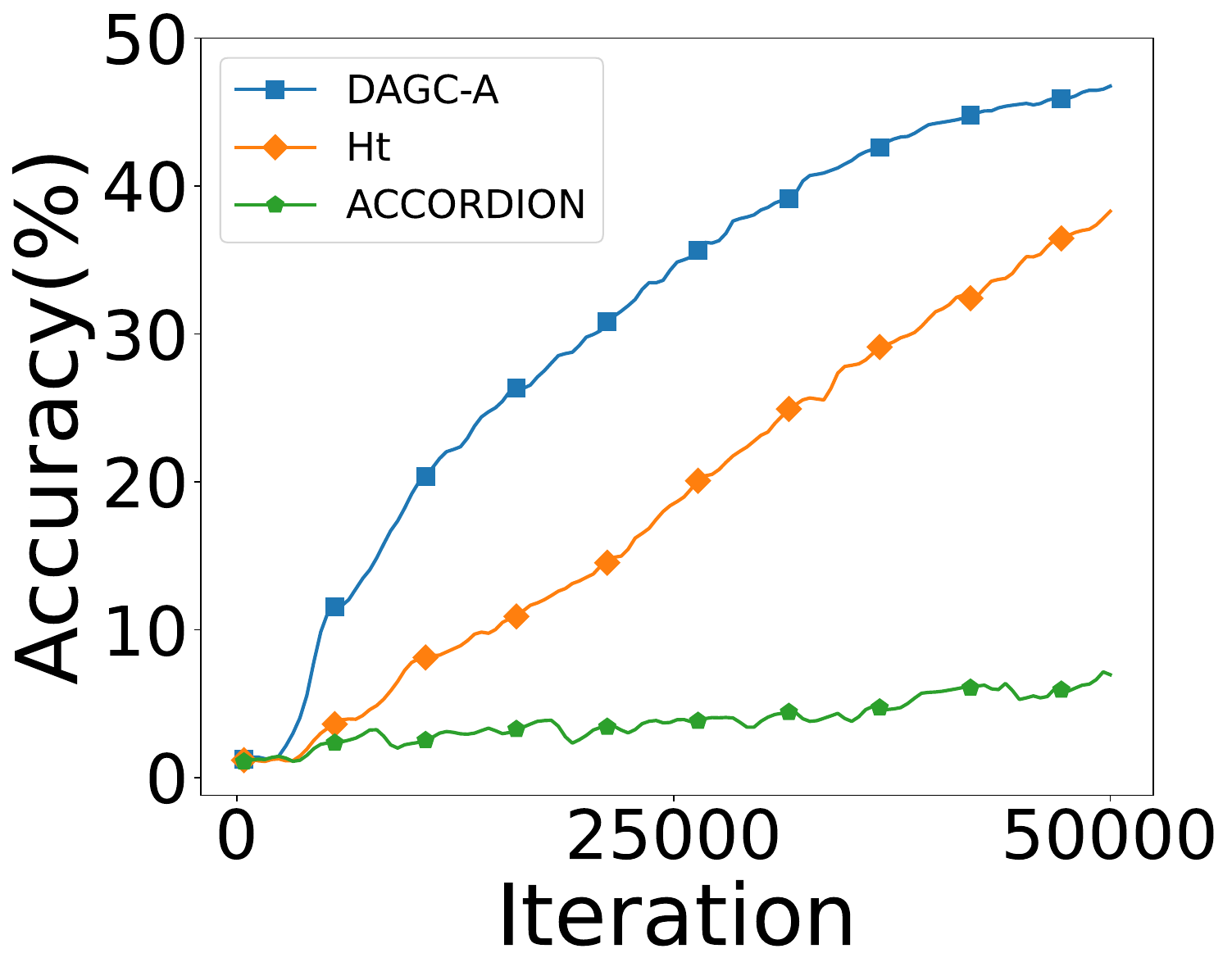}\label{fig:5:d}}
\caption{The training curves (Accuracy vs. Iterations) for ResNet18 @CIFAR-10 and VGG11@CIFAR-100 under the relative compression (a, c) and the absolute compression (b, d). DAGC performs better in all cases.}
\label{fig:5}
\end{figure}

\subsection{Comparsion of real time cost}\label{VI-F}

In Fig.~\ref{fig:4}, we show the training curves that illustrate accuracy over time for DAGC-R compared to other compression strategies and the baseline without compression. The experiments were conducted in the dynamic network, with an average bandwidth of 8.4 Mb/s \cite{zhang2021serverless}.

Fig.~\ref{fig:4:a} shows that at $\Bar{\delta}=0.1\%$, DAGC-R significantly reduces the time needed to reach $60\%$ (as well as $70\%$) accuracy. Specifically, it saves $16.78\%$, $17.56\%$ and $57.16\%$ ($12.75\%$, $11.12\%$ and $56.14\%$) compared to Top-$k$, ACCORDION, and training without compression, respectively. Similarly, Fig.~\ref{fig:4:b} reveals that, at $\bar{\lambda} = 0.05$, DAGC-A saves $36.19\%$, $50.64\%$ and $68.28\%$ (as well as $29.08\%$, $49.48\%$ and $71.47\%$) of the time required to achieve $60\%$ ($70\%$) accuracy compared to Ht, ACCORDION and training without compression.

Overall, DAGC-R demonstrates more efficient convergence, achieving high accuracy within a shorter time frame compared to both the baseline and other compression strategies.

\subsection{Scalability for models and datasets}\label{VI-G}

We expand the experiments to ResNet18 and VGG11 models, as well as the CIFAR-100 dataset, with the results shown in Fig.~\ref{fig:5}. The experiments uses $\text{SR}=100$, $\bar{\delta}=0.1\%$  for the relative compression and $\bar{\lambda}=0.05$ for the absolute compression.  In all sub-figures, DAGC outperforms the uniform compression, demonstrating that DAGC has excellent scalability for different models and datasets.

In Fig.~\ref{fig:5:a} (as well as Fig.~\ref{fig:5:b}),  DAGC-R (DAGC-A) can save up to $9.96\%$ ($52.74\%$) iterations compared to Top-$k$ (Ht). Similarly, in Fig.~\ref{fig:5:c} (as well as Fig.~\ref{fig:5:d}),  DAGC-R (DAGC-A) can save up to $16.67\%$ ($59.5\%$) iterations compared to Top-$k$ (Ht). It should be noted that Ht and ACCORDION using Ht have severe accuracy degradation, showing its bad scalability for large datasets and models.

\section{Related Works} \label{VII}

\noindent\textbf{Communication optimization in communication-constraint non-IID scenarios} aims to solve the communication bottleneck, while eliminating the accuracy degradation due to non-IID datasets. The work \cite{DGA} proposes the Delayed Gradient Averaging algorithm, which enhances the DML efficiency by delaying the gradient averaging step, allowing simultaneous local computation and communication. The work \cite{P-FedAVG} proposes a completely parallelizable FL algorithm, P-FedAvg, which extends the traditional FedAvg by allowing multiple parameter servers to collaboratively train a model, ensuring efficient convergence and scalability in a Parallel Federated Learning architecture. A Resource-efficient FL system is proposed in the work \cite{REFL} to tackle issues of resource heterogeneity in FL. There are also some works that propose to use a hierarchical architecture for DML in mobile environments, which matches the structure of LAN-WAN. The work \cite{HFL} first proposes a hierarchical architecture in communication-constrained non-IID scenarios. The work \cite{n-HFL} verifies the convergence of the hierarchical PS architecture in such scenarios. This work \cite{HFL-WAN-LAN} proposes a training scheduling strategy for DML in the LAN-WAN architecture. However, these methods neither mitigate the accuracy degradation brought about by non-IID scenarios nor are they friendly to mobile communication environments that charge by data usage. Compared to these methods, gradient compression algorithms save more communication costs savings in a per-traffic billing mobile network environment \cite{FedZIP, FedCOM}.

\noindent\textbf{Adaptive gradient compression} provides the ability to fine-tune compression parameters, a feature not often available in traditional algorithms  \cite{AdaQS,PASGD}. This enhances their robustness, particularly in varying scenarios that involve dynamic network environments and data heterogeneity. DC2 \cite{dc2}, a control setup based on network latency for handling compression, was proposed. This innovative system ensures timely completion of model training, even amidst fluctuating network conditions. This patent~\cite{SIDCOPATENT} introduces the system design of a statistical-based gradient compression method for a distributed training system that is based on the work~\cite{sidco}. The work \cite{cui2022optimal} proposes a systematic examination of the trade-off between compression and model accuracy in Federated Learning, introducing an adaptation framework to optimize the compression rate in each iteration for improved model accuracy while reducing network traffic. The work proposes a transmission strategy, called FedLC \cite{FedLC}, which combines model compression, forward error correction, and retransmission to improve the network utilization in FL with lossy communication.
The study introduces SkewScout \cite{niid2020icml}, aiming to enhance the robustness of algorithms particularly in non-IID scenarios. SkewScout achieves this by dynamically modulating the compression ratio according to the disparity in loss among workers, which is a parameter notoriously challenging to evaluate. As a result, the implementation of SkewScout becomes a complex task.

\noindent\textbf{Theoretically optimizing gradient compression:} Some works \cite{QuantityGlobal} attempt to improve the convergence rate, \textit{i.e.}, tighter upper bounds and lower time complexity, to optimize existing compression algorithms for mobile scenarios with communication constraints and heterogeneous data. The work \cite{smoothness-aware-gc,smoothness-aware-QSGD} utilizes smoothness matrices to boost existing compressors in both theory and practice. The study takes a comparison between D-QSGD and D-EF-SGD in non-IID conditions, integrating bias correction \cite{stich2020communication} to enhance their data dependency. The paper also provides a theoretical analysis of the robustness of the hard threshold sparsification algorithm which transmits solely the absolute gradient values exceeding a fixed hard threshold \cite{hardthreshold}. The findings indicate that compared to Top-$k$, this algorithm is more resilient when dealing with non-IID challenges. 

\noindent\textbf{Data-aware methods:} There have been studies that put forward the idea of data-aware node selection in DML. They introduce a technique \cite{amazon} that uses data volumes as a criterion for worker selection within Federated Learning. This stands in contrast to the traditional gradient-based methods \cite{wu2022node}. These experiments confirm that the data-volume-oriented node selection approach is superior to the uniform selection tactic in non-IID situations. This indicates that allowing  \textit{large workers} to transmit more data could potentially be advantageous. Yet there don't seem to be any existing studies introducing data-aware algorithms in the gradient compression domain. In our study, we introduce a data-aware gradient compression algorithm, accompanied by an in-depth theoretical analysis.
\section{Conclusion} \label{conclusion}
In the study, we introduce an innovative gradient compression algorithm, drawing from a fresh perspective by considering the unevenness in data volume sizes, which enhances its robustness with non-IID datasets. As an initial step, we present empirical evidence supporting the idea that assigning higher compression ratios to workers dealing with larger data volumes can accelerate convergence. Following this, we establish the convergence rate for non-uniform D-EF-SGD when applied in conjunction with either relative or absolute compressors. We derive the key factors, which greatly affect the model convergence in the communication-constrained non-IID environment. By minimizing this factor, we propose DAGC-R, which sets $\frac{\delta_i}{\delta_j}\approx(\frac{p_i}{p_j})^{2/3}$, and DAGC-A, where $(\frac{\lambda_i}{\lambda_j})^{2/3}=\frac{p_j}{p_i}$.
We assess the effectiveness of DAGC by conducting tests on both real-world datasets and artificially partitioned non-IID datasets. The results of these evaluation experiments showcase that DAGC has the capability to reduce the number of iterations by as much as $25.43\%$ on the real-world non-IID datasets. Furthermore, on artificially separated non-IID datasets, it enhances the accuracy by a substantial $3.14\%$.

\renewcommand*{\bibfont}{\footnotesize}

\printbibliography
\section{Appendix} \label{Appendix}
\subsection{Proof of \textbf{Theorem 1, 2, 3 }}\label{IV-F}
We define a virtual sequence that aids in our derivation, referring to \cite{stich2020communication}:
\begin{equation}
\tilde{\textbf{x}}_0 = \textbf{x}_0, \quad \tilde{\textbf{x}}_{t+1} := \tilde{\textbf{x}}_{t}-\gamma\sum_{i=1}^n p_i g_t^i \nonumber.
\end{equation}

The error term that  illustrates the gap between the virtual sequence and the actual sequence is represented as
\begin{equation}
 \tilde{\textbf{x}}_{t} - \textbf{x}_{t} = \gamma \sum_{i=1}^n p_i \textbf{e}_t^i \nonumber.
\end{equation}

For further discussion, let's define $G_t:=\textbf{E}\lVert \nabla f(x_t)\rVert^2$, $E_t=\sum_{i=1}^n p_i^2 \mathbb{E}\lVert \textbf{e}_t^i \rVert^2$, $\tilde{F}_t:=\textbf{E} f(\tilde{\textbf{x}}_t) - f^* $ and $F_t:=\textbf{E} f(\textbf{x}_t) - f^* $. 

\begin{lemma}
Considering a function $f$, which is $L$-smooth. If the learning rate $\gamma$ is less than or equal to $\frac{1}{4L}$,  the following is true for the iterations of non-uniform D-EF-SGD with relative compression:
\begin{equation}
\label{lemma:1a}
    \tilde{F}_{t+1}\leq \tilde{F}_t - \frac{\gamma}{4}G_t + \gamma^2 \frac{L \sum_{i=1}^n p_i^2\sigma^2}{2} + \gamma^3 \frac{n L^2}{2} E_t.
\end{equation}

If $f$ is exhibits $\mu$-convexity, the following is observed

\begin{equation}
\label{lemma:1b}
    X_{t+1}\leq (1-\frac{\gamma \mu}{2})X_t - \frac{\gamma}{2}F_t + \gamma^2 \sum_{i=1}^n p_i^2\sigma^2 + 3 \gamma^3 n L E_t.
\end{equation}
\end{lemma}

\textit{Proof.} Similar to the analysis in  \cite{stich2020communication}, we conclude 
\begin{eqnarray}
 \tilde{F}_{t+1}&\leq& \tilde{F}_t - \frac{\gamma}{4}G_t + \gamma^2 \frac{L}{2} \mathbb{E}\lVert \sum_{i=1}^n p_i \xi _t^i\rVert^2 
 \nonumber\\
 &+& \gamma^3 \frac{L^2}{2} \mathbb{E}\lVert \tilde{\textbf{x}}_t - \textbf{x}_t \rVert^2 \nonumber, \\
 X_{t+1}&\leq& (1-\frac{\gamma \mu}{2})X_t - \frac{\gamma}{2}F_t  + \gamma^2 \mathbb{E}\lVert \sum_{i=1}^n p_i \xi _t^i\rVert^2 \nonumber\\
 &+& 3 \gamma^3 L \mathbb{E}\lVert \tilde{\textbf{x}}_t - \textbf{x}_t \rVert^2 \nonumber.
\end{eqnarray}

With the independent $\xi_t^i$ and \textbf{Assumption 3} in \cite{stich2020communication}, the following equation emerges:
\begin{equation}
    \mathbb{E}_{\xi_t}\lVert \sum_{i=1}^n p_i \xi _t^i\rVert^2 = \sum_{i=1}^n p_i^2 \mathbb{E}_{\xi_t}\lVert\xi _t^i\rVert^2 \leq  \sum_{i=1}^n p_i^2 \sigma^2.  \nonumber
\end{equation}

Additionally, we derive:
\begin{equation*}
\mathbb{E}\lVert \tilde{\textbf{x}}_t - \textbf{x}_t \rVert^2 = \mathbb{E}\lVert \sum_{i=1}^n p_i e_t^i \rVert^2 \leq\footnote{The inequality follows from the fact that 
$\lVert \sum_{i=1}^k a_i \rVert^2 \leq k \sum_{i=1}^k \lVert a_i\rVert^2$.} n\sum_{i=1}^n p_i^2\mathbb{E}\lVert e_t^i \rVert^2 = n E_t.
\end{equation*}

Consequently, our targeted outcomes are achieved.

\begin{lemma}
It holds 

\begin{eqnarray}
E_{t+1} &\leq& (1-\frac{\delta_{min}}{2})E_t+\sum_{i=1}^n p_i^2 \sigma ^2 \nonumber \\
&+& \frac{2}{\delta_{min}} (C_{\zeta}\zeta^2+C_{Z} Z^2 G_t),
\label{lemma:2a}
\end{eqnarray}
where $C_{\zeta} = C_{Z} = \sum_{i=1}^n\frac{\delta_{min}}{\delta_i}p_i^2$.
\end{lemma}

\textit{Proof.} With the analysis in  \cite{stich2020communication}, it follows 

\begin{equation}
\begin{split}
    &\mathbb{E}_{\xi_t^i,C_\delta}\lVert \textbf{e}_{t+1}^i\rVert^2 
    \\&\leq (1-\frac{\delta}{2})\lVert \textbf{e}_{t}^i\rVert^2+\frac{2}{\delta}\lVert\nabla f_i(\textbf{x}_t)\rVert^2+(1-\delta)\sigma^2 \\
    &\leq (1-\frac{\delta}{2})\lVert \textbf{e}_{t}^i\rVert^2+\frac{2}{\delta}(\zeta_i^2+Z^2\lVert\nabla f(\textbf{x}_t)\rVert^2)+\sigma^2.
\label{lemma:2b}
\end{split}
\end{equation}

The final inequality is based on \textbf{Assumption 4} as cited in \cite{stich2020communication}. Then we incorporate the distinct compression ratios from various workers into Eq. \ref{lemma:2b} and aggregate the outcomes:
\begin{equation}
\begin{split}
    E_{t+1} &\leq (1-\frac{\delta_{min}}{2})\sum_{i=1}^n p_i^2 \lVert \textbf{e}_{t}^i\rVert^2\\&+\frac{2}{\delta_{min}}(\sum_{i=1}^n\frac{\delta_{min}}{\delta_i}p_i^2) (\zeta ^2+Z^2 G_t)+\sum_{i=1}^n p_i^2 \sigma^2, \nonumber
\end{split}
\end{equation}
where $\delta_{min} = \min\{\delta_1,\ldots,\delta_n\}, \zeta=\max\{\zeta_1,\ldots,\zeta_n\}$.
    
\begin{lemma}[Lyapunov function]
Considering a function $f$, which is $L$-smooth. If the learning rate $\gamma$ is less than or equal to $\dfrac{\delta_{min}}{4L Z \sqrt{n C_Z }}$. Then it holds

\begin{eqnarray}
\label{lemma:3a}
\Xi_{t+1}&\leq& \Xi_{t} - \frac{\gamma}{8}G_t + \gamma ^2 \frac{L \sum_{i=1}^n p_i^2 \delta ^2}{2}+ \nonumber\\
&&\gamma ^3 \left(\frac{L^2 n  }{\delta_{min}}\right)\left(\frac{2 C_{\zeta} \zeta^2}{\delta_{min}}+\sum_{i=1}^n p_i^2\sigma^2\right),
\end{eqnarray}
where $\Xi_t:= \tilde{F}_t+b E_t, b= \dfrac{\gamma^3 L^2 n}{\delta_{min}}$. Additionally, assuming $f$ is both $L$-smooth and $\mu$-convex and $\gamma$ is less or equal to $\dfrac{\delta_{min}}{14 L Z \sqrt{ n C_{Z}}}$, the following holds:

\begin{eqnarray}
\label{lemma:3b}
    \Psi_{t+1}&\leq& \left(1-\min\left\{\frac{\gamma \mu}{2},\frac{\delta}{4}\right\}\right)\Psi_{t} - \frac{1}{8L}G_t+\gamma ^2\sum_{i=1}^n p_i^2 \sigma ^2 \nonumber\\
    &&+ \gamma ^3\left(\frac{12 L n}{\delta_{min}}\right)\left (\frac{2 C_{\zeta} \zeta ^2}{\delta_{min}}+\sum_{i=1}^n p_i^2\sigma^2\right),
\end{eqnarray}
where $\Psi_t := X_t +a E_t$ with $a = \dfrac{12\gamma^3 n L}{\delta_{min}}$.
\end{lemma}

\textit{Proof.} For smooth functions, we  incorporate Eq.~\ref{lemma:1a} and ~\ref{lemma:2a} into the right side of the expression $\Xi_{t+1}:= \tilde{F}_{t+1}+b E_{t+1}$.

In the case of convex functions, we introduce Eq. \ref{lemma:1b} and  \ref{lemma:2a} into the right side of the expression $\Psi_{t+1} := X_{t+1} +a E_{t+1}$. This concludes the entire proof.

For the function that is non-convex, by integrating Eq. \ref{lemma:3a} with Appendix F's Lemma 27 from \cite{stich2020communication}, we successfully validate \textbf{Theorem \ref{theorem:1}}. When addressing a convex function where $\mu = 0$, by employing Eq. \ref{lemma:3b} in conjunction with Lemma 27 from \cite{stich2020communication} found in Appendix F, we confirm \textbf{Theorem \ref{theorem:2}}. For the function showcasing strong convexity characterized by $\mu > 0$, using Eq. \ref{lemma:3b} and referring to Lemma 25 in Appendix F of \cite{stich2020communication}, we establish the truth of \textbf{Theorem \ref{theorem:3}}.

\subsection{Proof of \textbf{Theorem \ref{theorem:4}}}\label{IV-G}

It's essential to recognize that the primary challenge can be converted to discerning the local optimal solution for the function $\Phi(\delta_1,\ldots,\delta_n)=\frac{\frac{p_1}{\sqrt{\delta_1}}+\ldots+\frac{p_n}{\sqrt{\delta_n}}}{\sqrt{\delta_{min}}}$  taking into account the constraint $\sum_{i=1}^n \delta_i = n\Bar{\delta}$ and the condition $\delta_i > 0$ for all values of $i$ ranging from $1$ to $n$. The demonstration of \textbf{Theorem \ref{theorem:4}} is broken down into two phases. In the initial phase, the problem with $n$
variables and a single constraint is recast into an optimization problem with only one variable, as depicted by Eq. \ref{lemma:4}. In the subsequent phase, the minimum for this single-variable optimization issue is determined. 

\begin{lemma}\label{lemma:4}

Suppose that $a_i,b_i >0, \forall i \in \{1,\ldots,n\}$ with $\sum_{i=1}^n a_i=A$ ($A$ is a constant), $b_i$ are constants, we have

\begin{equation}
    \sum_{i=1}^n \frac{b_i}{\sqrt{a_i}} \geq A^{-\frac{1}{2}}(\sum_{i=1}^n b_i^{\frac{2}{3}})^{\frac{3}{2}}.
\end{equation}

The inequality takes equal if $a_i= A b_i^{\frac{2}{3}}(\sum_{i=1}^n b_i^{\frac{2}{3}})^{-1}$.
\end{lemma}

\textit{Proof.} With the equality constrain on $a_i$, we define a Lagrangian function as follows:
\begin{equation}
\mathbb{L}=\sum_{i=1}^n \frac{b_i}{
\sqrt{a_i}}+\sigma (\sum_{i=1}^n a_i-A). \nonumber
\end{equation}

Based on the condition for optimality, we can deduce:
\begin{equation}
\left\{
             \begin{array}{ll}
             \frac{ \partial \mathbb{L} }{ \partial \sigma }&=\sum_{i=1}^n a_i-A=0\\
             \frac{ \partial \mathbb{L} }{ \partial a_i }&=-\frac{1}{2}b_i a_i^{-\frac{3}{2}}+\sigma = 0,  \forall i \in \{1,\ldots,n\} \nonumber
             \end{array}
\right. .
\end{equation}

From the system of equations provided, we can deduce the sought-after result.

With \textbf{Lemma~\ref{lemma:4}}, the function $\Phi(\delta_1,\ldots,\delta_n)$ is transformed into a one-dimensional function. Supposing $\delta_{min}=\delta_j \leq \min\{\delta_i\}, i\in\{1,\ldots,n\} \setminus \{j\}$, we define $b_i=p_i$ if $i\in[1,j-1]$, and for others $b_i=p_{i+1}$. We also set $a_i=\delta_i$ and $A=(n\Bar{\delta}-\delta_j)$.
\begin{equation}
    \Phi(\delta_1,\ldots,\delta_n) \geq \frac{p_j}{\delta_j} + \frac{(P-p_j^{\frac{2}{3}})^\frac{3}{2}}{\sqrt{(n\Bar{\delta}-\delta_j)\delta_j}},\,\text{where}\, P=\sum_{i=1}^n p_i^{\frac{2}{3}}.
\label{lemma:5a}
\end{equation}

Eq. \ref{lemma:5a}  is achieved  when $\delta_i= (n\Bar{\delta}-\delta_j) p_i^{\frac{2}{3}}(P-p_j^{\frac{2}{3}})^{-1}, i \neq j$. Given that $p_i$ is sorted in descending sequence, $\min\{\delta_i\}$ is $\delta_n$ if $j \in \{1,\ldots,n-1\}$, or else $\min\{\delta_i\}$ is $\delta_{n-1}$. 

Taking into account that the minimum of the right side of Eq. \ref{lemma:5a} depends on the range of $\delta_j$, we'll evaluate the scenarios for $j\in[1,n-1]$ and $j=n$ separately. 

\noindent$\bullet$ If $j \in [1,n-1]$, we have 
\begin{equation}
    \delta_{min}=\delta_j=\frac{(n\Bar{\delta}-\delta_j)p_n^{\frac{2}{3}}}{P-p_j^{\frac{2}{3}}}. \nonumber
\end{equation}

By setting $Q_j=\frac{P-p_j^{2/3}}{p_n^{2/3}},j\in[1,n-1]$ and using $\delta_j \leq \min\{\delta_i\}$. We deduce the range for $\delta_j \in (0,\frac{n\Bar{\delta}}{Q_j+1}]$. By defining $H(\delta_j)=\frac{p_j}{\delta_j}+\frac{(P-p_j^{\frac{2}{3}})^\frac{3}{2}}{\sqrt{(n\Bar{\delta}-\delta_j)\delta_j}}$, we can compute the derivative of $H(\delta_j)$:

\begin{equation}
\begin{split}
    H'(\delta_j)=&-p_j \delta_j^{-2}\\
    &-\frac{1}{2}(P-p_j^{\frac{2}{3}})^{\frac{3}{2}}[(n\Bar{\delta}-\delta_j)\delta_j]^{-\frac{3}{2}}(n\Bar{\delta}-2\delta_j) < 0. \nonumber
\end{split}
\end{equation}

Thus we get the minimum of $H(\delta_j)$ at $\delta_j=\frac{n\Bar{\delta}}{Q_j+1}$:

\begin{equation}
    H(\delta_j)\geq H(\frac{n\Bar{\delta}}{Q_j+1})=\frac{1}{n\Bar{\delta}}(p_n Q_j(1+Q_j)+p_j(1+Q_j)).
\label{lemma:5b}
\end{equation}

We combine Eq. \ref{lemma:5a} and \ref{lemma:5b} and complete the first case ($j\neq n$) in the proof.

\noindent$\bullet$ If $j=n$, we set $Q_n=\frac{P-p_n^{2/3}}{p_{n-1}^{2/3}}$ and have 

\begin{equation}
    \min\{\delta_i\}= \delta_{n-1}=\frac{n\Bar{\delta}-\delta_n}{Q_n}. \nonumber
\end{equation}

We get the range of $\delta_n$ is $(0,\dfrac{n\Bar{\delta}}{Q_n+1} ]$. In this range, $H'(\delta_j)<0$ (the proof process is the same as $j \neq n$). We have 

\begin{equation}
\begin{split}
    H(\delta_j)&\geq H(\frac{n\Bar{\delta}}{Q_n+1} )\\&=\frac{1}{n\Bar{\delta}}(p_n(1+Q_n)+p_{n-1} Q_n(1+Q_n)).
\end{split}
\label{lemma:5c}
\end{equation}

We combine Eq. \ref{lemma:5a} and \ref{lemma:5c} and complete the second case ($j=n$) in the proof.
\subsection{Proof of \textbf{Theorem \ref{non-convex D-EF-SGD-A}, \ref{convex D-EF-SGD-A}, \ref{strong convex D-EF-SGD-A}}}\label{V-E}

\begin{lemma}
Let $f$ be $L$-smooth. If the stepsize $\gamma \leq \frac{1}{4L}$, then it holds for the iterates of non-uniform D-EF-SGD with the absolute compressor with the absolute compressor:
\begin{equation}
\label{D-EF-SGD-A:a}
    \tilde{F}_{t+1}\leq \tilde{F}_t - \frac{\gamma}{4}G_t + \gamma^2 \frac{L \sum_{i=1}^n p_i^2\sigma^2}{2} + \gamma^3 \frac{n L^2 d}{2}\sum_{i=1}^n p_i^2 \lambda_i^2 .
\end{equation}

If $f$ is in addition $\mu$-convex, we have

\begin{equation}
\begin{split}
\label{D-EF-SGD-A:b}
    X_{t+1}&\leq (1-\frac{\gamma \mu}{2})X_t - \frac{\gamma}{2}F_t \\& + \gamma^2 \sum_{i=1}^n p_i^2\sigma^2 + 3 \gamma^3 n L d \sum_{i=1}^n p_i^2 \lambda_i^2.
\end{split}
\end{equation}

\end{lemma}

\textit{Proof.} According to the property of the absolute compressor, we have 
\begin{equation}
    E_t \leq  d \sum_{i=1}^n p_i \lambda_i^2.
\end{equation}

We combine the property and Lemma 1 and complete the proof.

For functions that is non-convexity, by incorporating Eq. \ref{D-EF-SGD-A:a} with Lemma 27 from Appendix F of \cite{stich2020communication}, we can validate \textbf{Theorem \ref{non-convex D-EF-SGD-A}}. When dealing with convex functions where $\mu = 0$, integrating Eq. \ref{D-EF-SGD-A:b} with Lemma 27 from Appendix F of \cite{stich2020communication} allows us to substantiate \textbf{Theorem \ref{convex D-EF-SGD-A}}. For functions exhibiting pronounced convexity (where$\mu > 0$), by merging Eq. \ref{D-EF-SGD-A:b} with Lemma 25 from Appendix F of \cite{stich2020communication}, we prove \textbf{Theorem \ref{strong convex D-EF-SGD-A}}.

\subsection{Proof of \textbf{Theorem \ref{theorem-DAGC-A}}}\label{V-F}

Note that the condition is that the total communication traffic is constrained. Previous work \cite{hardthreshold} demonstrated a conversion formula from the threshold $\lambda_i$ to the relative compression ratio $\delta_i$ in IID scenarios, \textit{i,e,.} $\lambda_i = \frac{D}{\sqrt{\delta_i}}$. This formula can not apply to this work, for that we focus on communication-constrained non-IID scenarios. 

To get the optimal $\lambda_i$, we assume that $\frac{\delta_i}{\delta_j} = (\frac{\lambda_i}{\lambda_j})^\Gamma$. Here, $\Gamma < 0$ since $\delta_i$ is negatively correlated to $\gamma_i$. Then we have $\sum_{i=1}^n \lambda_i^\Gamma = n \Bar{\lambda}^\Gamma$ according to $\sum_{i=1}^n \delta_i = n \Bar{\delta}$ in Sec.~\ref{IV-G}.

We use the Lagrange multiplier method and define a Lagrangian function as follows:
 
\begin{equation}
\mathbb{L}=\sum_{i=1}^n p_i^2 \lambda_i^2 + \sigma (\sum_{i=1}^n \lambda_i^\Gamma - n\Bar{\lambda}^\Gamma). \nonumber
\end{equation}

By the optimality condition, we have 
\begin{equation}
\left\{
             \begin{array}{ll}
             \frac{ \partial \mathbb{L} }{ \partial \lambda_i }&= 2p_i^2\lambda_i + \sigma \Gamma \lambda_i^{\Gamma-1} = 0,  \forall i \in \{1,\ldots,n\}\\
             \frac{ \partial \mathbb{L} }{ \partial \sigma }&=\sum_{i=1}^n \lambda_i^\Gamma - n\Bar{\lambda}^\Gamma=0 \nonumber
             \end{array}
\right. .
\end{equation}

According to $\frac{ \partial \mathbb{L} }{ \partial \lambda_i } = 0$, we have  $\frac{p_i}{p_j} = (\frac{\lambda_i}{\lambda_j})^{\frac{\Gamma}{2}-1}$. We combine this result with the property of DAGC-R, \textit{i.e.}, $\frac{\delta_i}{\delta_j}\approx(\frac{p_i}{p_j})^{2/3}$, resulting in $\Gamma \approx -1$. Which means $\lambda_i \propto \frac{1}{\delta_i}$  and $\frac{p_i}{p_j} \approx (\frac{\lambda_i}{\lambda_j})^{-\frac{3}{2}}$.

For ease of calculation, we next use the equation, \textit{i.e.}, $\frac{p_i}{p_j} = (\frac{\lambda_i}{\lambda_j})^{-\frac{3}{2}}$, rather than approximately equal. In this way, we have

\begin{equation}
\left\{
             \begin{array}{ll}
             2p_i^2\lambda_i - \sigma \lambda_i^{-2} &= 0,  \forall i \in \{1,\ldots,n\}\\
             \sum_{i=1}^n \lambda_i^{-1} - n\Bar{\lambda}^{-1} &= 0 \nonumber
             \end{array}
\right. .
\end{equation}

By solving system of equations above, the proof completes.

\vfill

\end{document}